\documentclass[sort&compress,final,3p,times]{elsarticle}
\usepackage[latin1]{inputenc}
\usepackage{amssymb}
\usepackage{amsmath}
\usepackage{amsthm}
\usepackage{esvect}
\usepackage{graphicx}
\usepackage{algorithmicx}
\usepackage{algorithm}
\usepackage{algpseudocode}
\usepackage{color}
\usepackage[table,xcdraw]{xcolor}
\usepackage{booktabs}
\usepackage{rotating,multirow,multicol}
\usepackage{makecell}
\usepackage{array}
\usepackage{tablefootnote}
\usepackage{blindtext}
\usepackage{pifont}
\usepackage{url} 
\usepackage[singlelinecheck=false]{caption}
\usepackage{subfig}
\usepackage{stackengine}
\usepackage[verbose]{placeins}
\usepackage{float}
\usepackage{lipsum}
\usepackage{color}
\usepackage{enumitem}
\usepackage{longtable}
\usepackage{cuted}
\usepackage{lineno}
\stripsep.sp

\usepackage{tikz}
\usetikzlibrary{tikzmark,decorations.pathreplacing,calc}

\newtheorem{proposition}{Proposition}
\newproof{pf}{Proof}
\newdefinition{defin}{Definition}
\newtheorem{lemma}{Lemma}
\newtheorem{assumption}{Assumption}

\newcommand{\hl}[1]{\setlength{\fboxsep}{1pt}\colorbox{lightgray}{#1}}

\DeclareMathOperator*{\argmin}{arg\,min}

\newcounter{remark}
\newenvironment{remark}[1]{\refstepcounter{remark}\medskip \textit{Remark~\theremark~-- #1}.}{\hfill\qedsymbol\medskip}

\algnewcommand{\LeftComment}[1]{\Statex \(\triangleright\) #1}

\algnewcommand{\GoTo}{\textbf{go to}~}%

\algblock{ParFor}{EndParFor}
\algnewcommand\algorithmicparfor{\textbf{parfor}}
\algnewcommand\algorithmicpardo{\textbf{do}}
\algnewcommand\algorithmicendparfor{\textbf{end\ parfor}}
\algrenewtext{ParFor}[1]{\algorithmicparfor\ #1\ \algorithmicpardo}
\algrenewtext{EndParFor}{\algorithmicendparfor}

\definecolor{TabR1}{HTML}{dce1e8}
\definecolor{TabR2}{HTML}{f2f5f9}

\let\today\relax
\makeatletter
\def\ps@pprintTitle{%
    \let\@oddhead\@empty
    \let\@evenhead\@empty
    \def\@oddfoot{\footnotesize\itshape
         { } \hfill\today}%
    \let\@evenfoot\@oddfoot
    }
\makeatother

\begin{document}
\begin{frontmatter}

\title{\emph{AFAFed}---Protocol analysis}

\address[label1]{Department of Information Engineering, Electronics and Telecommunications (DIET),\\Sapienza University of Rome, Via Eudossiana 18, 00184 Rome, Italy\vspace{0.5em}}

\cortext[cor1]{Corresponding author. Phone: +39 06 44585869, Fax: +39 06 44585632.}

\author[label1]{Enzo~Baccarelli}
\ead{enzo.baccarelli@uniroma1.it}
\author[label1]{Michele~Scarpiniti\corref{cor1}}
\ead{michele.scarpiniti@uniroma1.it}
\author[label1]{Alireza~Momenzadeh}
\ead{alireza.momenzadeh@uniroma1.it}
\author[label1]{Sima~Sarv~Ahrabi}
\ead{sima.sarvahrabi@uniroma1.it}
%


\begin{abstract}
In this paper, we design, analyze the convergence properties and address the implementation aspects of \textit{AFAFed}. This is a novel \textit{A}synchronous \textit{F}air \textit{A}daptive \textit{Fed}erated learning framework for stream-oriented IoT application environments, which are featured by time-varying operating conditions, heterogeneous resource-limited devices (i.e., coworkers), non-i.i.d. local training data and unreliable communication links. The \textit{key new} of \textit{AFAFed} is the synergic co-design of: (i) two sets of adaptively tuned tolerance thresholds and fairness coefficients at the coworkers and central server, respectively; and, (ii) a distributed adaptive mechanism, which allows each coworker to adaptively tune own communication rate. The convergence properties of \textit{AFAFed} under (possibly) non-convex loss functions is guaranteed by a set of \textit{new} analytical bounds, which formally unveil the impact on the resulting \textit{AFAFed} convergence rate of a number of Federated Learning (FL) parameters, like, first and second moments of the per-coworker number of consecutive model updates, data skewness, communication packet-loss probability, and maximum/minimum values of the (adaptively tuned) mixing coefficient used for model aggregation. 
\end{abstract}


\end{frontmatter}

\section{Motivations and goals}
\label{sec:introduction}

The incoming booming phase of the Internet of Things (IoT) era is characterized by an ever more increasing massive utilization of ever-smaller resource-constrained devices for the personalized and distributed support of ever more complex Machine Learning (ML)-aided stream wireless applications \cite{cisco1,Hanes2017}. Since IoT devices operate mainly at the network edge and connect to the backbone Internet through bandwidth-limited and failure-prone wireless (possibly, mobile) access links, the questions of \textit{where} and \textit{how} to process the gathered IoT big data are still open challenges \cite{Hanes2017}.

About \textit{where} to perform the processing, in the state-of-the-art cloud-centric approach, streams of data gathered by IoT devices at the network edge are uploaded to a remote centralized cloud-data center by exploiting the long-range Internet backbone \cite{Hanes2017}. However, we note that: (i) due to ever more stringent user privacy concerns, the recent worldwide legislation limits data storage only to what is user-consented and strictly necessary for processing \cite{Custers2019}; (ii) the centralized cloud-centric approach suffers from large and unpredictable network latencies \cite{cisco1}; and, (iii) massive data uploaded to the remote cloud cause severe traffic congestion in the Internet backbone \cite{Li2018a}. Hence, being the sources of IoT data remotely located outside the cloud, Fog Computing (FC) \cite{Baccarelli2017a} is emerging as a practical paradigm for the distributed support of ML applications, in which IoT devices and local servers are networked \cite{Baccarelli2019}, to bring data processing closer to where data is produced \cite{Baccarelli2021}.

Passing to consider \textit{how} data processing should be performed, we remark that, in order to guarantee data privacy while simultaneously enforce distributed training of complex ML models by resource-limited IoT devices, a decentralized paradigm for ML training called Federated Learning (FL) has been quite recently introduced \cite{McMahan2016}. In FC-supported FL systems, IoT devices (hereafter referred to as coworkers) employ own local data to cooperatively train an ML model shared with a central server hosted by a proximal Fog node. For this purpose, from time to time, coworkers upload their updated local ML models to the Fog-hosted central server for aggregation. The updating-aggregation steps are repeated in multiple rounds until a target accuracy of the trained ML model is achieved \cite{McMahan2016}.

Overall, the FL paradigm enables the privacy-preserving distributed training of complex ML models on resource-limited coworkers by operating only at the edge network \cite{Brisimi2018,Baccarelli2017,Jin2017}. However, several open issues must be still adequately addressed before implementing FL at scale in IoT stream realms, namely \cite{Mohammadi2018}: (i) unreliable communication; (ii) heterogeneous IoT devices; (iii) non-independent and identically distributed (non-i.i.d.) local data sets; (iv) fairness-related issues; (v) synchronization issues; and, (vi) time-varying operating environments. The above issues are inter-related, mainly because: (a) under device heterogeneity, an asynchronous approach to the FL could be appealing, in order to speed-up global convergence and avoid the slowdown induced by the slower coworkers (generally referred to as stragglers) \cite{Chen2019}. However, under non-i.i.d. local data sets, the resulting global FL solution attained at the convergence could be unfair, i.e., it could be too biased toward the local solutions of the faster coworkers. Then, suitable fairness mechanisms should be designed for coping with the asynchronous access of the faster coworkers to the central server; (b) in order to capitalize the faster convergence promoted by the asynchronous FL paradigm, each coworker should locally process fresh data without experiencing stalling phenomena. However, under stream applications, the arrival times of new data at the coworkers are random and (typically) out-of-control \cite{Mohammadi2018}. This requires, in turn, the design of suitable pro-active local data management policies, which account for the asynchronous behaviour of the coworkers and allow them to not stall their data processing; and, (c) under wireless realms, coworker-to-central server links are typically unreliable and heterogeneous, i.e., they exhibit non-vanishing packet loss rates that may vary from coworker to coworker \cite{Hanes2017}. This may be a potential additional source of unfairness, because the global FL solution could be biased towards the local solutions of the coworkers that experience more reliable links. Hence, it demands, in turn, for the joint design of transmission scheduling policies at the coworkers and fairness-aware aggregation mechanisms at the central server that account also for link unreliability.

Overall, since the afforded challenges are inter-related, a unified formal framework is required for capturing their inter-dependency and, then, a joint approach should be developed for their integrated solution.

\paragraph{Reference scenario}

\begin{figure*}[htb]
\centering
\includegraphics[width=0.60\textwidth]{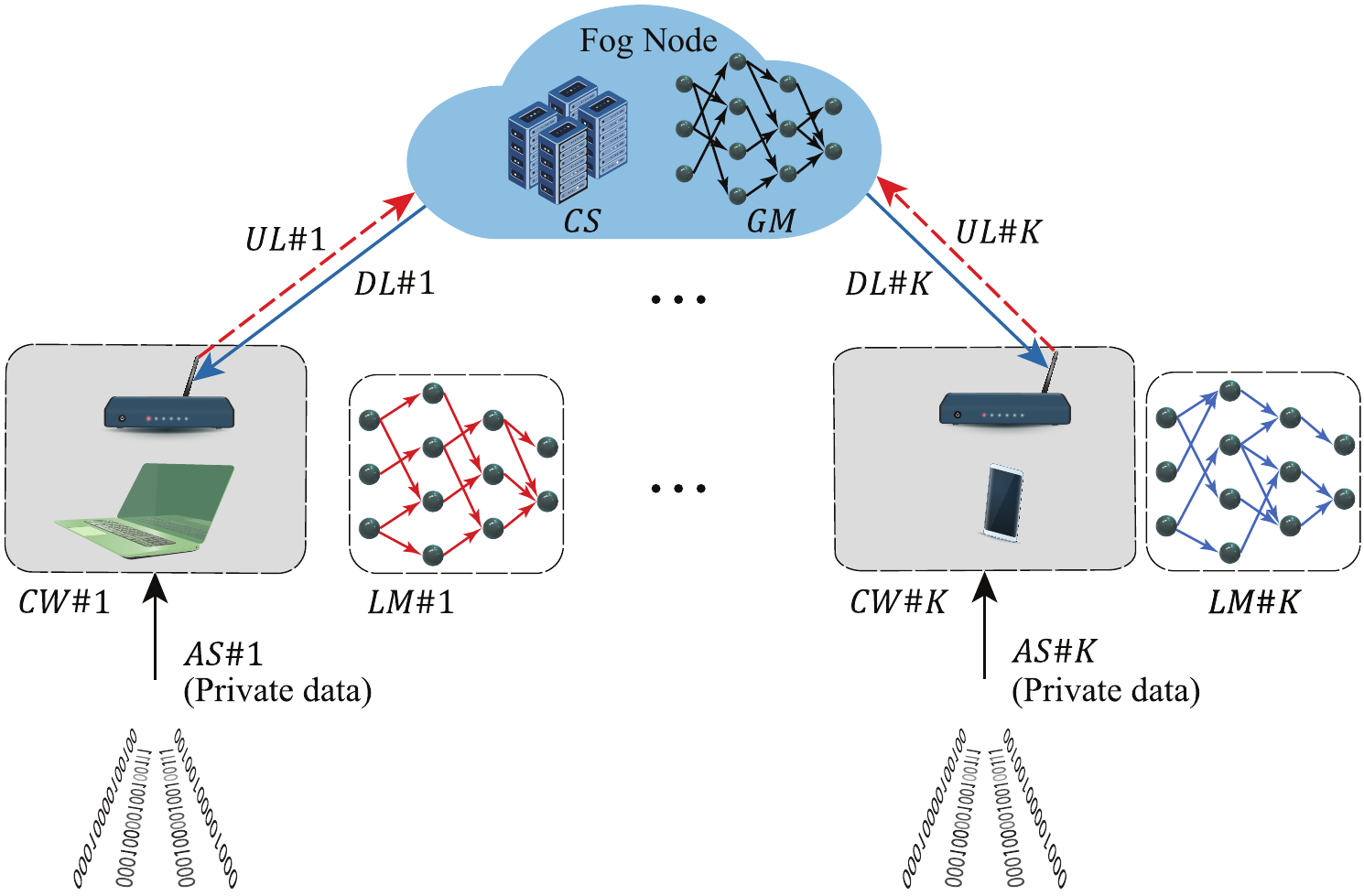}
\caption{The considered \textit{AFAFed} reference scenario. Continuous (dashed) arrows indicate reliable (unreliable) communication links. CW:= CoWorker; K:= Number of involved coworkers; AS:= Arrival Stream; LM:= Local ML Model; CS:= Central Server; GM:= Global ML Model; UL:= Up-Link; DL:= Down-Link.}
\label{fig:scenario}
\end{figure*}

Motivated by these considerations, we consider the reference scenario of Fig.\ \ref{fig:scenario}, in which: 
\begin{enumerate}
	\item the trained ML model is \textit{general}, i.e. it may range from a simple Linear Regressor (LR), a binary Support Vector Machine (SVM) to a (complex) multi-class Deep Neural Network (DNN) \cite{Goodfellow2016}. Furthermore, depending on the underlying ML model, the performed training may be \textit{supervised}/\textit{unsupervised} or a mix of them \cite{Goodfellow2016}. The target loss function to minimize through the FL may be \textit{non-convex};
	\item the involved coworkers may be equipped with \textit{heterogeneous} computing/communication/storage local resources, and their local training data may be non-i.i.d. Furthermore, the arrivals of new data at the coworkers of Fig.\ \ref{fig:scenario} may be \textit{non-synchronized}, and they may happen at random times;
	\item coworkers perform multiple consecutive local model updates before sending their own current local models to the central server in an \textit{asynchronous} way. The number of the performed consecutive local updates may be \textit{time-varying} and \textit{different} from coworker to coworker;
	\item coworker-to-server uplinks in Fig.\ \ref{fig:scenario} may be \textit{unreliable} (i.e., affected by communication delays, packet-losses, and/or intermittent failure phenomena); and,
	\item after receiving an updated model update from a coworker, the central server aggregates it into the global model \textit{without waiting} for additional updates from other coworkers and, then, returns the updated global model to the sending coworker. The age (i.e., staleness) of each local updating received by the central server is modeled as a random variable (r.v.) with \textit{unbounded} support.
\end{enumerate}

\paragraph{Paper contributions}
By referring to the depicted scenario, the focus of this paper is on the: (a) optimized algorithmic design; (b) convergence analysis; and, (c) protocol design of \textit{AFAFed}---an adaptive asynchronous scheme for performing FL in stream realms that are supported by networked FC technological platforms. We expect that the main \textit{key news} that characterize the \textit{AFAFed} proposal are as follows:
\begin{enumerate}
	\item in order to cope with the (aforementioned) \textit{non-i.i.d.} nature of the local training data, \textit{AFAFed} implements a \textit{distributed} mechanism, which is inspired by the formal tool of the so called Imperfect Consensus \cite{McMahan2017}. Its main new is that it is based on the \textit{adaptive} setting of some suitable tolerance thresholds and allows each coworker to develop a \textit{personalized} local model that accounts for the statistical heterogeneity of the available local training data. For this purpose, \textit{AFAFed} implements also a novel \textit{distributed} mechanism for the \textit{adaptive} tuning of the number of the per-coworker inter-aggregation local model updates, which aims to reduce the per-coworker transmission rates (i.e., save bandwidth) while accounting for the (randomly distributed) arrivals of new data. At this regard, in order to cope with the (typically, unpredictable) random fluctuations of the inter-arrival times of the data streams of Fig. 1, the \textit{AFAFed} coworkers implement an (\textit{ad-hoc} designed) distributed and adaptive policy for the management of their \textit{finite-size} buffers, which allows the coworkers to process data at their maximum processing speeds, so to speed up the resulting \textit{AFAFed} convergence time;
	\item in order to cope with the \textit{resource-heterogeneity} of the coworkers and their related fairness issues, \textit{AFAFed} implements a novel mechanism for the \textit{adaptive} tuning of some suitably defined \textit{fairness coefficients}. The goal is to build up a \textit{fair} global model at the central server that treats all the participating coworkers in an egalitarian way. For this purpose, \textit{AFAFed} implements a novel adaptive \textit{age} and \textit{fairness-aware asynchronous} aggregation mechanism at the central server, in which each received local model is aggregated by weighting it by an \textit{adaptively tuned} coefficient that \textit{jointly} accounts for its age \textit{and} fairness level; and, finally, 
	\item we derive a set of \textit{analytical bounds}, which guarantee the asymptotic convergence (in expectation) of the global model built up by the central server to a neighborhood of a stationary point of the underlying (possibly, nonconvex) loss function. Since the numerical evaluations of the proposed bounds rely on some system parameters that could be \emph{a priori} unknown, we also design a companion protocol for their \textit{online profiling}. 
\end{enumerate}

\paragraph{Adopted notation}
We point out that $\vv{w}$ indicates a column vector, $|\mathcal{S}|$ is the size (i.e., the cardinality) of the set $\mathcal{S}$, $\left\lceil \cdot \right\rceil$ (resp., $\left\lfloor \cdot \right\rfloor$) is the ceil (resp., floor) function, while $\triangleq$ means ``equal by definition''. Furthermore, $\xi \sim \mathcal{D}$ indicates that the r.v. $\xi$ follows the probability distribution $\mathcal{D}$, and $E_\mathcal{D}\left\{\cdot\right\}$ (resp., $E_{X|Y}\left\{\cdot\right\}$) denotes the expectation with respect to (w.r.t.) the probability distribution $\mathcal{D}$ (resp., w.r.t. the distribution of the r.v. $X$ conditioned to $Y$), while $E\left\{\cdot\right\}$ means joint expectation w.r.t. all the r.v.'s involved.

\section{Asynchronous FL: Preliminaries and Reference Models}
\label{sec:Asynchronous_FL}

The main goal of this section is to formally introduce the \textit{AFAFed} learning problem and the adopted solving approach. The main taxonomy of the paper and the definition of the main adopted notation is recapped in Appendix \ref{appendix:A}.

\subsection{Local and global loss functions}
\label{sec:local_global}

ML models for statistical inference require that a set of model parameter $\vv{w} \in \mathbb{R}^l$ is learnt (possibly online) on the basis of (possibly time varying streams of distributed) training data. Hence, let:
\begin{equation}
S_k \triangleq \left\{ \left( \vv{x}_i \in \mathbb{R}^d, \, \vv{y}_i \in \mathbb{R}^{\mathcal{C}}, \; 1 \leq i \leq \left| S_k \right| \right) \right\}, \quad k = 1, \ldots, K,
\label{eq:S_k}
\end{equation}
be the set of (possibly time-varying) training data available at the $k$-th coworker (i.e. $coworker\#k$) of Fig.\ \ref{fig:scenario}. Under supervised training \cite{Goodfellow2016}, each data of the set in Eq. \eqref{eq:S_k} is composed of: (i) a feature vector $\vv{x}_i \in \mathbb{R}^d$, which acts as input of the ML model to be trained; and (ii) a vector $\vv{y}_i \in \mathbb{R}^{\mathcal{C}}$, which represents the $\mathcal{C}$-ary ground-truth label associated to the feature $\vv{x}_i$. After assuming $S_k \cap S_j = \emptyset$ for $k \neq j$, in FL stream application scenarios, the elements of $S_k$ may vary over the time but they may be accessed only by $coworker\#k$. Furthermore, in stationary non-i.i.d. environments, the probability distribution $\mathcal{D}_k$, $k = 1, \ldots, K$, of the training data set $S_k$ is time invariant, but it may depend on the coworker index $k$. 

After indicating by $\vv{w}_k \in \mathbb{R}^l$, $k = 1, \ldots, K$, the weight vector of the ML model estimated by $coworker\#k$, let: $\mathcal{T}: \mathbb{R}^d \times \mathbb{R}^l \rightarrow \mathbb{R}^{\mathcal{C}}$ be the $\vv{w}_k$-dependent input-output transformation implemented by the considered ML model. Hence, by design, the label $\widehat{\vv{y}}_i \in \mathbb{R}^{\mathcal{C}}$ generated by the ML model when its input is $\vv{x}_i \in S_k$ is formally defined as follows \cite{Goodfellow2016}: $\widehat{\vv{y}}_i \triangleq \mathcal{T}\left( \vv{x}_i, \vv{w}_k \right)$. The deviation of the estimated label from the ground-truth one is measured by an assigned (possibly non-convex but differentiable) scalar real-valued loss function, formally defined as follows \cite{Goodfellow2016}: $\ell\left( \vv{y}_i, \widehat{\vv{y}}_i, \vv{w}_k \right): \mathbb{R}^{\mathcal{C}} \times \mathbb{R}^{\mathcal{C}} \times \mathbb{R}^l \rightarrow \mathbb{R}$. In the case of ML models that require unsupervised learning (for example the K-means model), the corresponding loss function does not longer depend on the ground-truth label $\vv{y}_i$. Some examples of convex/non-convex loss functions for supervised/unsupervised training are reported, for example, in Table I of \cite{Wang2019}. 

Since the probability distribution $\mathcal{D}_k$ is typically \textit{a priori} unknown, in practical application scenarios, the expectation: $E_{\left( \vv{x}_i, \vv{y}_i \right) \sim \mathcal{D}_k}\left( \ell\left( \vv{y}_i, \widehat{\vv{y}}_i, \vv{w}_k \right) \right)$ of the loss function w.r.t. $\mathcal{D}_k$ is replaced by the corresponding $k$-th empirical risk \cite{Goodfellow2016}:
\begin{equation}
F_k\!\left( \vv{w}_k \right) \triangleq \frac{1}{\left| S_k \right|} \sum_{\left( \vv{x}_i, \vv{y}_i \right) \in S_k} \ell\left( \vv{y}_i, \widehat{\vv{y}}_i, \vv{w}_k \right),
\label{eq:F_k}
\end{equation}
whose gradient w.r.t the weight vector reads as in:
\begin{equation}
\vv{\nabla}F_k\!\left( \vv{w}_k \right) \triangleq \frac{1}{\left| S_k \right|} \sum_{\left( \vv{x}_i, \vv{y}_i \right) \in S_k} \vv{\nabla}_{\vv{w}} \ell\left( \vv{y}_i, \widehat{\vv{y}}_i, \vv{w}_k \right).
\label{eq:DF_k}
\end{equation}
Hence, according to the (aforementioned) paradigm of the \textit{Agnostic FL} \cite{Mohri2019}, the global empirical risk $F\left(\vv{w}\right)$ over all data sets of Eq. \eqref{eq:S_k} is: 
\begin{equation}
F\left(\vv{w}\right) \triangleq \sum_{k=1}^K \lambda_k F_k\left( \vv{w} \right),
\label{eq:F_global}
\end{equation}
where: $\left\{ \lambda_k \in \left[ 0, \, 1 \right], \, k = 1, \ldots, K \right\}$ is a set of fairness coefficients which sum up to the unit. These coefficients play the role of (possibly, \textit{a priori} unknown and/or time-varying) hyper-parameters, which must be suitably optimized, in order to guarantee the attainment of some target fairness level among the involved coworkers \cite{Mohri2019}.
According to the definition in Eq. \eqref{eq:F_global}, the resulting Jain's \textit{Fairness Index} ($FI$) for measuring the inter-coworker fairness reads as in:
\[
FI = \frac{\left( \sum_{k=1}^K \lambda_k \right)^2}{K \left( \sum_{k=1}^K \lambda_k^2 \right)}.
\]
In our setting, it ranges from $1/K$ (most unfair case) to the unit (perfect fairness), and it is maximum when all $\lambda$'s coefficients share a same value.

\subsection{Learning problem formulation and model personalization}
\label{sec:problem_formulation}

The goal of the learning problem is to minimize the global risk function in Eq. \eqref{eq:F_global}, i.e., to compute:
\begin{equation}
\vv{w}^\ast \triangleq \argmin_{\vv{w}} \left\{ F\left( \vv{w} \right) \right\},
\label{eq:w_star}
\end{equation}
so that,
\begin{equation}
F^\ast \triangleq F\left( \vv{w}^\ast \right) \equiv \min_{\vv{w}}\left\{ F\left( \vv{w} \right) \right\} > - \infty,
\label{eq:F_star}
\end{equation}
is the global minimum of $F(\cdot)$. The key challenge of carrying out the above minimization is that, from Eqs. \eqref{eq:F_k} and \eqref{eq:DF_k}, it follows that the computation of $F(\cdot)$ in Eq. \eqref{eq:F_global} and its gradient:
\begin{equation}
\vv{\nabla}F\!\left( \vv{w} \right) \triangleq \sum_{k=1}^K \lambda_k \vv{\nabla}F_k\!\left( \vv{w} \right),
\label{eq:DF}
\end{equation}
would require that the full set $S \triangleq \cup_{k=1}^K S_k$ of training data is available at the central server of Fig.\ \ref{fig:scenario}. Since this requirement contrasts, indeed, with the privacy preserving nature of the FL paradigm, we will pursue a solving approach that is inspired by the so-called \textit{Imperfect Consensus} paradigm for the distributed optimization of global objective function under proximity local constraints \cite{Koppel2017}. Shortly, we point that \textit{Imperfect Consensus} is a generalization of the original \textit{Consensus} paradigm \cite{Boyd2011}. This last one is a formal technique for turning an additive global objective function which does not split due to the presence of a global variable shared by the involved additive terms into a summation of separable local objectives, which are amenable to distributed local optimization. However, when the local training sets in Eq. \eqref{eq:S_k} are non-i.i.d., enforcing ``perfect'' consensus among the local models evaluated by the coworkers may degrade the resulting local (i.e., per-coworker) prediction accuracy \cite{Koppel2017}. This is the reason for introducing ``soft'' local proximity constraints, which enforce ``imperfect'' consensus by allowing local models to be close but not necessary equal. 

Motivated by these considerations, we formally introduce a set of per-coworker local variables: $\Big\{ \vv{w}_k \in \mathbb{R}^l, \, k = 1, \ldots, K \Big\}$ and a single global variable $\vv{\overline{w}} \in \mathbb{R}^l$ tied to the central server. Afterwards, we replace the centralized minimization problem in Eq. \eqref{eq:w_star} by the following distributed one:
%
\begin{subequations}\label{eq:distributed_minimization}
  \allowdisplaybreaks
	\begin{align}
	& \underset{\mbox{\normalsize{$\vv{w}_1, \ldots, \vv{w}_K, \vv{\overline{w}}$}}}{\mbox{\large{$\min$}}} \:\; \sum_{k=1}^K \lambda_k F_k\!\left( \vv{w}_k \right) \, ,\label{eq:distributed_minimization_1} \\[2ex]
	& \text{s.t.:} \quad \left\| \vv{w}_k - \vv{\overline{w}} \right\|^2 \leq B_k \, , \hphantom{======} \label{eq:distributed_minimization_2} 
	\end{align}
\end{subequations}
%
where $\left\{ B_k, \, k = 1, \ldots, K \right\}$ are non-negative scalar-valued tolerance thresholds. This last formulation is referred to as the Imperfect Consensus form of the learning problem in Eq. \eqref{eq:w_star}, since the set of constraints in Eq. \eqref{eq:distributed_minimization_2} forces all the local variables $\left\{ \vv{w}_k \right\}$ to be close enough to a same global variable $\vv{\overline{w}}$. Perfect consensus is attained for vanishing values of the tolerance thresholds at the r.h.s.'s of Eq. \eqref{eq:distributed_minimization_2}.

\begin{remark}{Model Personalization and adaptive per-coworker ball constraints}\label{remark:1}
A key reason for adopting the Imperfect Consensus-based formulation in Eqs. \eqref{eq:distributed_minimization_1} and \eqref{eq:distributed_minimization_2} is that, under non-i.i.d. FL scenarios, the sets of local proximity constraints in Eq. \eqref{eq:distributed_minimization_2} with non-vanishing tolerance thresholds $\left\{ B_k \right\}$ may be used for \textit{personalizing} the models $\left\{ \vv{w}_k \right\}$ locally evaluated by coworkers.
Several algorithms have been recently proposed to mitigate data heterogeneity by adding a local regularization term to the co-worker's objective functions, and popular examples of such algorithms include \textit{FedProx} in \cite{Li2020d} and \textit{FedAsync} in \cite{Xie2020}, to name a few. However, we argue that it would be worth to explore algorithms that add \textit{adaptive} ball constraints when minimizing local objectives, in order to \textit{adaptively} control the variability of local models and eventually attempt to align the steady-state values of global and local models when the radius $\left\{ B_k \right\}$ of the ball constraints vanish in the limit. In this regard, we anticipate that a peculiar \textit{and new} feature of the \textit{AFAFed} proposal is that the radius $\left\{ B_k \right\}$ of the ball constraints in Eq. \eqref{eq:distributed_minimization_2} are \textit{not} \textit{a priori} fixed but they are \textit{adaptively} tuned, in order to allow the implemented FL scheme to attain a right trade-off among the contrasting requirements of model personalization and model generalization. 
\end{remark}

Being the Imperfect Consensus learning problem in Eqs. \eqref{eq:distributed_minimization_1} and \eqref{eq:distributed_minimization_2} a constrained optimization problem, its solution may be characterized by resorting to the evaluation of the saddle-points of the corresponding Lagrangian function. This last one is defined as in \cite{bazaraa2017}: 
\begin{equation}
\mathcal{L} \equiv \mathcal{L}\!\left( \left\{ \vv{w}_k \right\}, \vv{w}_k \left\{ \mu_k \right\} \right) \triangleq \sum_{k=1}^K \lambda_k F_k\!\left( \vv{w}_k \right) + \frac{\mu_k}{2} \left( \left\| \vv{w}_k - \vv{\overline{w}} \right\|^2 - B_k \right),
\label{eq:Lagrangian}
\end{equation}
where: (i) $\left\{ \vv{w}_k \right\}$ and $\vv{\overline{w}}$ play the role of $(K+1)$ primal optimization variables; while, (ii) $\left\{ \mu_k \right\}$ are non-negative Lagrange multipliers associated to the proximity constraints in Eq. \eqref{eq:distributed_minimization_2} (i.e., dual variables). Let:
\begin{equation}
\left\{ \left\{ \vv{w}_k^\ast \right\}, \left\{ \mu_k^\ast \right\}, \vv{\overline{w}}^\ast \right\},
\label{eq:star_solution}
\end{equation}
be the (possibly, not unique) optimal values of the primal and dual variables which solve the constrained problem in Eq. \eqref{eq:distributed_minimization}. Hence, these optimal values must be a (possibly, not unique) saddle-point of the Lagrangian function in Eq. \eqref{eq:Lagrangian}, that is, they must be a (possibly, not unique) solution of the following max-min problem:
\begin{equation}
\max_{\vv{\mu} \geq \vv{0}} \left\{ \min_{\vv{w}_1, \ldots, \vv{w}_K, \vv{\overline{w}}} \: \mathcal{L}\!\left( \left\{ \vv{w}_k \right\}, \vv{w}_k \left\{ \mu_k \right\} \right) \right\}.
\label{eq:max_min_Lagrangian}
\end{equation}
This means, in turn, that the $\left( (K+1)l + K \right)$-dimensional vector: $\vv{\nabla}\mathcal{L}$ of the gradients of the Lagrangian function in Eq. \eqref{eq:Lagrangian} performed w.r.t. the primal and dual variables must vanish at the optimum, that is\footnote{In the case of non-convex global loss function $F(\cdot)$ in Eq. \eqref{eq:F_global}, the set of solutions of Eq. \eqref{eq:Lagrangian_vanishing} embraces local/global maxima, local/global minima and/or saddle-points of $F(\cdot)$ \cite{bazaraa2017}.}:
\begin{equation}
\vv{\nabla}\mathcal{L} = \vv{0}_{(K+1)l + K}.
\label{eq:Lagrangian_vanishing}
\end{equation}
According to Eq. \eqref{eq:Lagrangian}, the partial gradients of the Lagrangian function $\mathcal{L}(\cdot)$ w.r.t. the primal and dual variables read as follows:
%
\begin{subequations}\label{eq:partial_gradient}
  \allowdisplaybreaks
	\begin{align}
	\vv{\nabla}_{\vv{w}_k}\mathcal{L} &= \lambda_k \vv{\nabla}F_k\!\left( \vv{w}_k \right) + \mu_k \left( \vv{w}_k - \vv{\overline{w}} \right), \quad k = 1, \ldots, K,  \label{eq:partial_gradient_1} \\
	\vv{\nabla}_{\vv{\overline{w}}}\mathcal{L} &= \sum_{k=1}^K \mu_k \left( \vv{\overline{w}} - \vv{w}_k \right), \quad k = 1, \ldots, K,  \label{eq:partial_gradient_2} \\
	\intertext{and,}
	\vv{\nabla}_{\mu_k}\mathcal{L} &= \left( \left\| \vv{w}_k - \vv{\overline{w}} \right\|^2 - B_k \right), \quad k = 1, \ldots, K.  \label{eq:partial_gradient_3} 
	\end{align}
\end{subequations}

\subsection{Asynchronous distributed updating of local models}
\label{sec:distributed_updating}

In the \textit{AFAFed} framework, the solution of the distributed leaning problem in Eqs. \eqref{eq:distributed_minimization_1} and \eqref{eq:distributed_minimization_2} is numerically evaluated by implementing a set of distributed SGD iterations, where the coworkers of Fig.\ \ref{fig:scenario} proceed in a \textit{fully asynchronous} way. This means that each coworker and the central server are equipped with own local clocks which run without any synchronization. Formally speaking, the following assumption holds:
\begin{assumption}[\textit{The AFAFed asynchronous setting}]
We assume that:
\begin{enumerate}
	\item $coworker\#k$ is equipped with a non-negative integer-valued local time index $t^{(k)}$ that counts the number of performed local iterations, so that $\left\{ \vv{w}_k \left( t^{(k)} \right), \, t^{(k)} = 0, 1, \ldots \right\}$ is the corresponding $k$-th sequence of local model weights;
	\item the central server is equipped with a global non-negative integer-valued time index $t$ that counts the number of performed global updates, so that $\left\{ \vv{w}(t), \, t = 0, 1, \ldots \right\}$ is the corresponding sequence of global model weights;
	\item all the introduced time indexes evolve in an un-coordinated way, i.e., there is not present any master clock.
\end{enumerate}
\vspace{-1em}\hfill $\blacksquare$
\end{assumption}

When the volumes $\left\{\left| S_k \right|\right\}$ of local training data are large, it may be computationally challenging to compute the full gradients in Eq. \eqref{eq:DF_k}. In such cases, SGD is typically used \cite{Goodfellow2016}. It employs the gradients \cite{Goodfellow2016}: 
\begin{equation}
\vv{\nabla}\widetilde{F}_k\!\left( \vv{w}_k; t^{(k)} \right) \triangleq \frac{1}{\left| MB_k \right|} \sum_{\left( \vv{x}_i, \vv{y}_i \right) \in MB_k \left( t^{(k)} \right)} \vv{\nabla}_{\vv{w}} \ell\!\left( \vv{y}_i, \widehat{\vv{y}}_i, \vv{w}_k \right),
\label{eq:DF_tilde_k}
\end{equation}
of the partial loss functions:
\begin{equation}
\widetilde{F}_k\!\left( \vv{w}_k; t^{(k)} \right) \triangleq \frac{1}{\left| MB_k \right|} \sum_{\left( \vv{x}_i, \vv{y}_i \right) \in MB_k \left( t^{(k)} \right)} \ell\!\left( \vv{y}_i, \widehat{\vv{y}}_i, \vv{w}_k \right),
\label{eq:F_tilde_k}
\end{equation}
which are defined on the basis of time-varying randomly sampled mini-batches: $MB_k\left( t^{(k)} \right)$, $k = 1, \ldots, K$, of the local data. Hence, after replacing the full gradient $\vv{\nabla} F_k\!\left( \vv{w}_k \right)$ in Eq. \eqref{eq:partial_gradient_1} by its stochastic approximation $\vv{\nabla}\widetilde{F}_k\!\left( \vv{w}_k; t^{(k)} \right)$ in Eq. \eqref{eq:DF_tilde_k}, the SGD iterations on the $k$-th primal and dual variables $\vv{w}_k$ and $\mu_k$ carried out by $coworker\#k$ at the local time $t^{(k)}$ read as follows (see Eqs. \eqref{eq:partial_gradient_1} and \eqref{eq:partial_gradient_3}):
\begin{subequations}\label{eq:SGD_iteration}
  \allowdisplaybreaks
	\begin{align}
	\vv{w}_k \left( t^{(k)} + 1 \right) &= \vv{w}_k \left( t^{(k)} \right) - \eta_k^{(0)}\left( t^{(k)} \right) \left[ \lambda_k^{(LAST)} \vv{\nabla}\widetilde{F}_k\!\left( \vv{w}_k\left( t^{(k)} \right) \right) + \mu_k \left( t^{(k)} \right) \left( \vv{w}_k \left( t^{(k)} \right) - \vv{\overline{w}}_k^{(LAST)} \right) \right] ,  \label{eq:SGD_iteration_1} \\
	\intertext{and,}
	\mu_k \left( t^{(k)} + 1 \right)    &= \left[ \mu_k \left( t^{(k)} \right) + \eta_k^{(1)}\left( t^{(k)} \right) \left[ \left\| \vv{w}_k\left( t^{(k)} \right) - \vv{\overline{w}}_k^{(LAST)} \right\|^2  - B_k \right] \right]_+ , \quad t^{(k)} = 0, 1, \ldots  \label{eq:SGD_iteration_2} 
	\end{align}
\end{subequations}
In Eqs. \eqref{eq:SGD_iteration_1} and \eqref{eq:SGD_iteration_2}, we have that: (i) $[z]_+$ means: $\max\left\{0, z\right\}$; (ii) $\eta_k^{(0)} \left( t^{(k)} \right)$ and $\eta_k^{(1)} \left( t^{(k)} \right)$ are (possibly, time-varying) scalar step-sizes to be suitably designed for improving the convergence speed of the iterations; (iii) $\lambda_k^{(LAST)}$ is the \textit{most} recent value of the $k$-th fairness coefficient $\lambda_k$ in Eq. \eqref{eq:distributed_minimization_2} communicated by the central server to $coworker\#k$; and, (iv) $\vv{\overline{w}}_k^{(LAST)}$ is the most recent instance of the global weight vector $\vv{\overline{w}}$ that $coworker\#k$ received from the central server. In this regard, we point out that, due to the asynchronous setting considered here (see \textit{Assumption 1}), $\vv{\overline{w}}_k^{(LAST)}$ may depend on the coworker index $k$, i.e., at a same wall clock time, \textit{different} instances of $\vv{\overline{w}}^{(LAST)}$ may be locally available at different coworkers\cite{Zhang2020}.

\subsection{Unreliable uplink communication}
\label{sec:uplink_communication}

In the scenario of Fig.\ \ref{fig:scenario}, coworkers are equipped with resource-constrained (typically, battery-powered) devices, while the central sever is hosted by a fixed resource-rich Fog node. Hence, the coworker-to-central server uplinks are the bottleneck, since they are typically more power and bandwidth constrained than the corresponding central server-to-coworker downlinks \cite{Niknam2020}. According to this consideration, we introduce the following formal assumption:
\begin{assumption}[\textit{AFAFed communication modeling}]
By referring to Fig.\ \ref{fig:scenario}, we assume that:
\begin{enumerate}
	\item the downlinks are ideal, i.e., delay and packet-loss free;
	\item the $k$-th uplink may be affected by both communication delays and packet-loss phenomena, where:
	\begin{itemize}
		\item communication delays are featured by an i.i.d random sequence: $\left\{ T_k^{(COM)}\!\left( t^{(k)} \right), \, t^{(k)} = 0, 1, \ldots \right\}$, where $\;$ $T_k^{(COM)}\!\left( t^{(k)} \right) \triangleq b_k/R_k\!\left( t^{(k)} \right)$, with: (i) $b_k$ (bit) being the size of the model updated by $coworker\#k$; and, (ii) $R_k\!\left( t^{(k)} \right)$ (bit/link-use) being the corresponding randomly time-varying uplink transmission rate at $t^{(k)}$;
		\item packet loss phenomena are modeled by an i.i.d. binary random sequence: $\left\{ CH_k\!\left( t^{(k)} \right) \in \left\{ 0, \, 1 \right\},  t^{(k)} = 0, 1, \ldots \right\}$, where $CH_k\!\left( t^{(k)} \right)$ is zero when a packet loss happens at $t^{(k)}$.
	\end{itemize}
\end{enumerate}
\vspace{-1em}\hfill $\blacksquare$
\end{assumption}
\noindent According to this assumption, we denote by:
\begin{equation}
P_k^{(LOSS)} \triangleq Prob\left\{ CH_k = 0 \right\} \equiv 1 - E\left\{ CH_k \right\}, \quad k = 1, \ldots, K,
\label{eq:P_loss}
\end{equation}
the resulting steady-state packet-loss probability of the $k$-th uplink of Fig.\ \ref{fig:scenario}.

\subsection{Asynchronous model aggregation}
\label{sec:model_aggregation}

In principle, $coworker\#k$ could provide updates to the central server in form of Lagrangian gradients or model vectors. Although these two alternatives are equivalent under ideal communication, nevertheless, it has been experienced that model-based updates are more robust than the corresponding gradient-based ones against the impairing effects induced by communication packet-loss phenomena \cite{Yang2019,Peng2019}. Hence, in order to formally characterize the asynchronous access to the central server performed by the involved \textit{AFAFed} coworkers, let: $\left\{ \xi_k(t) \in \left\{ 0, \, 1 \right\}, \, k = 1, \ldots, K \right\}$ be the set of binary r.v.'s so defined:
%
\begin{subequations}\label{eq:csi}
  \allowdisplaybreaks
	\begin{align}
  &\xi_k \left(t\right) \triangleq
  \begin{cases}
	 1, & \text{if the central server is updated by the $k$-th coworker at global time $t$}, \\
   0, & \text{otherwise}
  \end{cases}  \label{eq:csi_k} \\
	\intertext{where the constraint:}
	&\sum_{k=1}^K \xi_k(t) = 1,  \label{eq:csi_k_constraint} 
	\end{align}
\end{subequations}
%
assures that, at each global index time $t$, the central server is updated by a single coworker. Then, by leveraging the gradient expression in Eq. \eqref{eq:partial_gradient_2} and the definition in Eq. \eqref{eq:csi_k}, the gradient descent update of the global model $\vv{\overline{w}}(\cdot)$ at the central server reads as in: 
\begin{equation}
\vv{\overline{w}}(t+1) = \vv{\overline{w}}(t) + \eta^{(2)}(t) \left[ \sum_{k=1}^K \xi_k(t+1) \left( \vv{w}_k(t+1) - \vv{\overline{w}}(t) \right) \right], \quad t = 0, 1, \ldots
\label{eq:CS_update_1}
\end{equation}
After exploiting the constraint in Eq. \eqref{eq:csi_k_constraint}, Eq. \eqref{eq:CS_update_1} can be rewritten, in turn, in the following two equivalent forms: 
%
\begin{subequations}\label{eq:CS_update_2}
  \allowdisplaybreaks
	\begin{align}
	\vv{\overline{w}}(t+1) &= \left( 1 - \beta(t) \right) \vv{\overline{w}}(t) + \beta(t) \left[ \sum_{k=1}^K \xi_k(t+1) \vv{w}_k(t+1) \right]  \label{eq:CS_update_2_1} \\
	   & \equiv \vv{\overline{w}}(t) - \beta(t) \,\widehat{\vv{G}}\!\left( \vv{\overline{w}}(t); \left\{ \xi_k(t+1) \right\} \right),  \label{eq:CS_update_2_2} 
	\end{align}
\end{subequations}
%
where:
\begin{equation}
\widehat{\vv{G}}\!\left( \vv{\overline{w}}(t); \left\{ \xi_k(t+1) \right\} \right) \triangleq \left[ \sum_{k=1}^K \xi_k(t+1) \left( \vv{\overline{w}}(t) - \vv{w}_k(t+1) \right) \right],
\label{eq:G}
\end{equation}
plays the role of a global stochastic gradient, and:
\begin{equation}
\beta(t) \equiv \beta\left( \left\{ \xi_k(t+1) \right\} \right) \in [0, \, 1],
\label{eq:beta}
\end{equation}
is a (possibly time varying) non-negative \textit{mixing} parameter, which is limited to the unit and may also depend on the current values of the binary access r.v.'s in Eq. \eqref{eq:csi_k}. Interestingly enough, the companion relationships in Eqs. \eqref{eq:CS_update_2_1} and \eqref{eq:CS_update_2_2} point out that the asynchronous model aggregation performed by the central server may be viewed under the \textit{equivalent} forms of moving average or SGD iterates.

\section{The \textit{AFAFed} proposal--Algorithms and Protocols}
\label{sec:AFAFed_proposal}

The modeling carried out in Section \ref{sec:Asynchronous_FL} points out that, in the \textit{AFAFed} framework: 
\begin{enumerate}
	\item the variables $\left\{ \vv{w}_k \right\}$ and $\vv{\overline{w}}$ (resp., $\left\{ \mu_k \right\}$) are the primal (resp., dual) \textit{optimization variables} of the learning problem in Eqs. \eqref{eq:distributed_minimization_1} and \eqref{eq:distributed_minimization_2}; 
	\item the fairness coefficients $\left\{ \lambda_k \right\}$ play the role of (possibly, time varying) \textit{hyper-parameters}, which must be \textit{suitably designed}, in order to throttle the global model $\vv{\overline{w}}$ towards a suitable fair solution; 
	\item the (possibly time varying) tolerance thresholds $\left\{ B_k \right\}$ in Eq. \eqref{eq:distributed_minimization_2} are non-negative hyper-parameters, to be designed for attaining a right personalization-vs.-generalization trade-off; 
	\item the (possibly time varying) sets of step sizes $\left\{ \eta_k^{(0)} \right\}$ and $\left\{ \eta_k^{(1)} \right\}$ in Eqs. \eqref{eq:SGD_iteration_1} and \eqref{eq:SGD_iteration_2} must be suitably designed, to speed up the convergence of the primal and dual iterates carried out by the coworkers; and,
	\item the (possibly time varying) mixing parameter $\beta \left(\cdot\right)$ in Eqs. \eqref{eq:CS_update_2_1} and \eqref{eq:CS_update_2_2} is a \textit{hyper-parameter} to be designed for performing fair and age-aware model aggregation.
\end{enumerate}

The above remarks fix the main directions pursued for the optimized design of the \textit{AFAFed} protocol.

\subsection{Overall view of the \textit{AFAFed} protocol}
\label{sec:AFAFed_protocol}

Algorithm \ref{alg:AFAFed_protocol} provides an overall view of the protocol that implements \textit{AFAFed}. It is composed of two main segments, i.e. a local segment to be run in parallel by coworkers (see the rows of Algorithm \ref{alg:AFAFed_protocol} numbered in azure), and a global segment to be run by the central server in a sequential way (see the rows of Algorithm \ref{alg:AFAFed_protocol} numbered in green). These two protocol segments will be  described in the sequel. In Algorithm  \ref{alg:AFAFed_protocol}, gray-shadowed equation numbering highlights the main algorithmic news of the \textit{AFAFed} proposal.

\begin{algorithm*}[htbp]
\caption{--- Overall view of the proposed \textit{AFAFed} protocol}
\label{alg:AFAFed_protocol}
\footnotesize
\vspace*{1pt}
\textbf{Input:} 
\vspace*{-4pt}
{\begin{itemize}
\setlength{\itemsep}{0pt}
\setlength{\parskip}{0pt}
\item Number $T$ of global iterations to be carried out by the central server;
\item Number $K$ of the involved coworkers.
\end{itemize}}
\textbf{Output:} 
\vspace*{-4pt}
{\begin{itemize}
\setlength{\itemsep}{0pt}
\setlength{\parskip}{0pt}
\item Set of the fairness coefficients $\left\{ \lambda_k(T) \right\}$;
\item Set of the weights vectors $\left\{ \vv{w}_k(T) \right\}$ of the local models;
\item Weight vector $\vv{\overline{w}}(T)$ of the global model.
\end{itemize}}
\begin{algorithmic}[1]
\LeftComment{\textit{Initialization phase}}
\State Set $\vv{\overline{w}}$ and $\left\{ \vv{w}_k \right\}$ to a same initial value $\vv{w}_0$;
\State Set all $\lambda_k$'s to $1/K$.
\vspace*{4pt}
\LeftComment{\textit{Sequential iterative phase at the central server}}
\For{$ t = 0 : \left( T - 1 \right) $}
\vspace*{4pt}
\LeftComment{\textit{Parallel iterative phase at the coworkers}}
\ParFor {$k = 1:K$}
\algrenewcommand{\alglinenumber}[1]{\color{blue}\footnotesize#1:}
\State {$coworker\#k$ runs $Iter_k$ primal/dual local iterations in Eqs. \eqref{eq:SGD_iteration_1} and \eqref{eq:SGD_iteration_2} by updating the step-sizes as in Eqs. \hl{\eqref{eq:eta_update_1}} and \hl{\eqref{eq:eta_update_2}}};
\State {$coworker\#k$ updates its local clock index as: $t^{(k)} := t^{(k)} + Iter_k$};
\State {$coworker\#k$ updates its tolerance threshold $B_k$ as in Eq. \hl{\eqref{eq:B_k}}};
\State {$coworker\#k$ updates its averaged Lagrange multiplier $\overline{\mu}_k$ as in Eq. \hl{\eqref{eq:mu_average}}};
\State {$coworker\#k$ sends the 3--ple: $\left\{ \vv{w}_k, \, \overline{\mu}_k, \, timestamp_k \right\}$ to the central server and starts its $Timer_k$};
\State {$coworker\#k$ updates the number of local iterations $Iter_k$ to be carried out in the next batch as in Eq. \hl{\eqref{eq:Iter_k}}};
\If {($Timer_k$ expired) \& (no feedback is received from the central server)}
\State \GoTo {Step\#5};
\EndIf \Comment{End-if at Step\#11}
\If {a new arrival happens at the $coworker\#k$}
\State {$coworker\#k$ applies the buffer management policy of \hl{Section \ref{sec:AFAFed_throughput}}};
\EndIf \Comment{End-if at Step\#14}
\Statex {}
\algrenewcommand{\alglinenumber}[1]{\color{green!80!black}\footnotesize#1:}
\If {central server receives an update from $coworker\#k$}
\State {central server updates the global upper and lower fairness thresholds $TH_U$ and $TH_L$ as in Eqs. \hl{\eqref{eq:TH_U}} and \hl{\eqref{eq:TH_L}}};
\State {central server updates the global up/down scaling factors $\Psi$ and $V$ as in Eq. \hl{\eqref{eq:example_up_down_func}}};
\If {$\left( \overline{\mu}_k > TH_U \right)$}
\State {central server updates $\lambda_k$ as in Eq. \hl{\eqref{eq:lambda_scale_up}} and normalizes the overall set $\left\{ \lambda_j \right\}$ as in Eq. \eqref{eq:lambda_new}};
\ElsIf { $\left( \overline{\mu}_k < TH_L \right)$}
\State {central server updates $\lambda_k$ as in Eq. \hl{\eqref{eq:lambda_scale_down}} and normalizes the overall set $\left\{ \lambda_j \right\}$ as in Eq. \eqref{eq:lambda_new}};
\EndIf \Comment{End-if at Step\#20}
\State {central server updates the model $age \left(\cdot\right)$ and mixing parameter $\beta \left(\cdot\right)$ as in Eqs. \eqref{eq:age} and \hl{\eqref{eq:beta_t}}, respectively};
\State {central server updates the global model weight vector $\vv{\overline{w}}$ as in Eq. \eqref{eq:CS_update_2_1}};
\State {central server broadcasts the updated set $\left\{ \lambda_j^{(NEW)}, \, j = 1,\ldots, K \right\}$ of the fairness coefficients to \textit{all} coworkers};
\State {central server sends a timestamped copy of the updated model $\vv{\overline{w}}$ to $coworker\#k$ and updates its global time index as: $t := t+1$};
\EndIf \Comment{End-if at Step\#17}
\Statex {}
\algrenewcommand{\alglinenumber}[1]{\color{blue}\footnotesize#1:}
\If {$coworker\#k$ receives a new timestamped global model $\vv{\overline{w}}$ from the central server}
\State {$coworker\#k$ updates its $timestamp_k$ and set: $\vv{w}_k^{(LAST)} := \vv{\overline{w}}$ in Eqs. \eqref{eq:SGD_iteration_1} and \eqref{eq:SGD_iteration_2}};
\State {$coworker\#k$ restarts its primal/dual iterations in Eqs. \eqref{eq:SGD_iteration_1} and \eqref{eq:SGD_iteration_2} by setting: $\vv{w}_k\left( t^{(k)} \right) := \vv{\overline{w}}$};
\State \GoTo {Step\#5};
\EndIf \Comment{End-if at Step\#30}
\Statex {}
\If {$coworker\#k$ receives an updated fairness coefficient $\lambda_k^{(NEW)}$ from the central server}
\State {$coworker\#k$ sets $\lambda_k^{(LAST)} := \lambda_k^{(NEW)}$ in Eq. \hl{\eqref{eq:SGD_iteration_1}}};
\EndIf \Comment{End-if at Step\#35}
\Statex {}
\algrenewcommand{\alglinenumber}[1]{\color{black}\footnotesize#1:}
\EndParFor \Comment{End parallel-for at Step\#4}
\EndFor \Comment{End for at Step\#3}
\Statex {}
\State \textbf{return} $\left\{ \lambda_k(T) \right\}$, $\left\{ \vv{w}_k(T) \right\}$, and $\vv{\overline{w}}(T)$.
\end{algorithmic}
\end{algorithm*}

\normalsize

\subsection{\textit{AFAFed} procedures at the co-workers}
\label{sec:AFAFed_procedures}

In the \textit{AFAFed} framework, each $coworker\#k$ maintains and updates a local model vector $\vv{w}_k$ (see $line\#5$ of Algorithm \ref{alg:AFAFed_protocol}), while a main goal of the central server is to update the corresponding global model vector $\vv{\overline{w}}$ (see $line\#26$ of Algorithm \ref{alg:AFAFed_protocol}). All these local and global models are initialized to a same value in the bootstrap phase (see $line\#1$ of Algorithm \ref{alg:AFAFed_protocol}). The distributed nature of the \textit{AFAFed} framework is featured by the {\sffamily\bfseries parallel-for} cycle in $line\#4$, which forces all the embraced steps to be executed in parallel by the coworkers. At the same time, its asynchronous nature stems from the fact that there is not a centralized master clock for aligning the local and global time indexes in $line\#6$ and $line\#28$ of Algorithm \ref{alg:AFAFed_protocol}.

\paragraph{Distributed and adaptive tuning of the Lagrange multipliers}

Unlike previous contributions \cite{Li2020a,Xie2020,Zhang2015a,Chen2020}, a main (and, indeed, \textit{new}) feature of the \textit{AFAFed} proposal is that each coworker is equipped with its own Lagrange multiplier $\mu_k$, which is time-varying and is locally updated in $line\#5$ of Algorithm \ref{alg:AFAFed_protocol} according to the iterations of Eq. \eqref{eq:SGD_iteration_2}. In the \textit{AFAFed} ecosystem, its role is, indeed, twofold. First, it provides an adaptive and personalized (i.e., per-coworker) measure of the penalization to be imposed to the deviation of the current local model $\vv{w}_k$ from the last received global one: $\vv{\overline{w}}_k^{(LAST)}$ (see Eq. \eqref{eq:SGD_iteration_2}). Second, it enforces imperfect consensus by adaptively throttling $\vv{w}_k$ to be close to (but not necessarily coincident with) the last received version of the global model $\vv{\overline{w}}_k^{(LAST)}$ (see Eq. \eqref{eq:SGD_iteration_1}). In order to smooth the oscillatory phenomena typically induced by too abrupt environmental changes, a second pursued design choice is to utilize the corresponding per-coworker sample average:
\begin{equation}
\overline{\mu}_k\left( t^{(k)} + 1 \right) \triangleq \frac{1}{t^{(k)} + 2} \sum_{i=0}^{t^{(k)} + 1} \mu_k(i), \quad t^{(k)} = 0, 1, \ldots,
\label{eq:mu_average}
\end{equation}
for the adaptive tuning of the fairness coefficients. For this purpose, $coworker\#k$ locally updates the sample average in Eq. \eqref{eq:mu_average} at each local time index $t^{(k)}$, and from time to time, communicates it to the central server for further processing (see $line\#8$ of Algorithm \ref{alg:AFAFed_protocol}).

\paragraph{Distributed and adaptive tuning of the number of local iterations}

Each $coworker\#k$ may perform a cluster of $Iter_k \geq 1$ local iterations between two consecutive communications to the central server. A \textit{new} feature of the \textit{AFAFed}  proposal is that the current value of the sample average in Eq. \eqref{eq:mu_average} is used as scaling coefficient, in order to make both \textit{distributed} and \textit{adaptive} the updating of $Iter_k$ in $line\#10$ of Algorithm \ref{alg:AFAFed_protocol}. Intuitively, the pursued design principle is to increase (resp., decrease) $Iter_k$ when the current value of $\overline{\mu}_k$ in Eq. \eqref{eq:mu_average} is low (resp., high). The rationale is that, since $\overline{\mu}_k$ is an (averaged) measure of the penalty to impose to the current deviation of the local model $\vv{w}_k$ from the global one $\vv{\overline{w}}_k^{(LAST)}$, it may be, indeed, reasonable to lower the communication coworker-to-central server communication rate (that is, to increase the number of performed consecutive local iterations) when the local model is sufficiently aligned with the global one. In order to formally implement this design principle, let:
\begin{equation}
\Omega_k \left( \overline{\mu}_k \right) : \mathbb{R}_0^+ \longrightarrow [1, \, + \infty ),
\label{eq:Omega_k}
\end{equation}
be a (possibly, $k$-dependent) non-negative function of the averaged Lagrange multiplier $\overline{\mu}_k$ in Eq. \eqref{eq:mu_average}, which: (i) is not decreasing for increasing $\overline{\mu}_k$; and, (ii) is unit valued in the origin. Hence, we design to adaptively tune $Iter_k$ as follows:
\begin{equation} 
Iter_k\left( t^{(k)} + 1 \right) = \max\left\{ 1; \, \left\lceil \frac{Iter_k^{(MAX)}}{\Omega_k \left( \overline{\mu}_k \left(t^{\left(k\right)}+1\right) \right)} \right\rceil \right\}, \quad t^{(k)} = 0, 1, \ldots,
\label{eq:Iter_k}
\end{equation}
where the positive integer-valued constant $Iter_k^{(MAX)}$ fixes the maximum number of consecutive local iterations allowed $coworker\#k$. 
Before proceeding, two explicative remarks are in order. First, an eligible example of $\Omega_k$-function is the following \textit{max-log} one:
\begin{equation}
\Omega_k \left( \overline{\mu}_k \right) = \max\left\{ 1; \, \min\left\{ Iter_k^{(MAX)}; \, a^{c \times \overline{\mu}_k} \right\} \right\}, \quad \text{with} \; a > 1, \; \text{and} \; c > 0.
\label{eq:Omega_max_log}
\end{equation}
Second, from a statistical point of view, $Iter_k$ in Eq. \eqref{eq:Iter_k} plays the role of an r.v., which assumes values over the integer-valued set: $\left\{ 1, 2, \ldots, Iter_k^{(MAX)} \right\}$. According to the relationship in \eqref{eq:Iter_k}, the probability density function (pdf) of $Iter_k$ depends on the pdf of the r.v. $\overline{\mu}_k$ in Eq. \eqref{eq:mu_average}, which, in turn, may be \textit{a priori} unknown. In this regard, we anticipate that the issue of the online profiling of the statistics of $Iter_k$ will be addressed in Section \ref{sec:AFAFed_implementation_aspects}.

\paragraph{Distributed and adaptive tuning of step-sizes}

The adaptive design of the $\eta$'s step-sizes in Eqs. \eqref{eq:SGD_iteration_1} and \eqref{eq:SGD_iteration_2} is driven by the following two considerations. First, in order to allow both fast responsiveness in the transient-state and stable behavior in the steady-state, it is likelihood that $\eta_k^{(0)}$ and $\eta_k^{(1)}$ scale up with the modules: $\left\| \vv{\nabla}\widetilde{F}_k \right\|$ and $\left\| \left\| \vv{w}_k\!\left( t^{(k)} \right) - \vv{\overline{w}}_k^{(LAST)} \right\|^2 - B_k\!\left( t^{(k)} \right) \right\|$ of the gradients of the involved Lagrangian function. This setting is supported, indeed, by the formal results reported in \textit{Theorem 3.3} of \cite{Peng2016} about the convergence of primal/dual iterations under nonstationary environments. Second, in order to attain a good personalization-vs.-generalization trade-off, it is reasonable to allow the step-sizes in Eq. \eqref{eq:SGD_iteration_1} and \eqref{eq:SGD_iteration_2} to scale up for increasing values of $\Omega_k$ in Eq. \eqref{eq:Omega_k}, so to increase the personalization (resp., generalization) capability of $coworker\#k$ when the current value of the corresponding averaged Lagrange multiplier $\overline{\mu}_k$ in Eq. \eqref{eq:mu_average} is large (resp., small). Overall, on the basis of these considerations, we \textit{design} the following two relationships for the per-iteration adaptive tuning of the step-sizes in Eqs. \eqref{eq:SGD_iteration_1} and \eqref{eq:SGD_iteration_2}: 
%
\begin{subequations}\label{eq:eta_update}
  \allowdisplaybreaks
	\begin{align}
	\eta_k^{(0)}\!\left( t^{(k)} \right) &= \max\left\{ \eta_{MIN}; \, \min\left\{ \eta_{MAX}; \, \Omega_k\!\left( t^{(k)} \right) \left\| \vv{\nabla}\widetilde{F}_k\!\left( t^{(k)} \right) \right\| \right\} \right\}, \quad t^{(k)} = 0, 1, \ldots,  \label{eq:eta_update_1} \\
	\intertext{and,}
	\eta_k^{(1)}\!\left( t^{(k)} \right) &= \max\left\{ \eta_{MIN}; \, \min\left\{ \eta_{MAX}; \, \Omega_k\!\left( t^{(k)} \right) \left\| \left\| \vv{w}_k\!\left( t^{(k)} \right) - \vv{\overline{w}}_k^{(LAST)} \right\|^2 - B_k\!\left( t^{(k)} \right) \right\| \right\} \right\}, \quad t^{(k)} = 0, 1, \ldots,  \label{eq:eta_update_2} 
	\end{align}
\end{subequations}
%
where $\eta_{MIN}$ (resp., $\eta_{MAX}$) is a non-negative lower (resp., upper) bound on the allowed step-size values, which acts as a regularization term \cite{Goodfellow2016}.

\paragraph{Distributed and adaptive tuning of the tolerance thresholds}

It is expected that the personalization (resp., generalization) capability of the local models $\left\{ \vv{w}_k \right\}$ (resp., of the global model $\vv{\overline{w}}$) increases for increasing (resp., decreasing) values of $\left\{ B_k \right\}$ in Eq. \eqref{eq:distributed_minimization_2}. This requires, in turn, that the overall set of these tolerance thresholds is \textit{jointly} optimized, in order to attain a good personalization-vs.-generalization trade-off. In this regard, a remarkable (and, at the best of the authors' knowledge, \textit{new}) design feature of \textit{AFAFed} is that even the values of the tolerance thresholds are \textit{tuned} in a \textit{distributed} and \textit{adaptive} way. Specifically, the pursued design criterion is allow $coworker\#k$ to autonomously increase (resp., decrease) the current value of $B_k\!\left( t^{(k} \right)$ when the corresponding average Lagrange multiplier $\overline{\mu}_k$ in Eq. \eqref{eq:mu_average} increases (resp., decreases), so to enhance model personalization (resp., model generalization). For this purpose, let:
\begin{equation}
\Theta\!\left( \overline{\mu}_k \right): \mathbb{R}_0^+ \longrightarrow \mathbb{R}_0^+,
\label{eq:Theta}
\end{equation}
be a non-negative non-decreasing function of $\overline{\mu}_k$, which vanishes in the origin. Hence, in $line\#7$ of Algorithm \ref{alg:AFAFed_protocol}, we design to adapt $B_k$ as follows:
\begin{equation}
B_k\!\left( t^{(k)} \right) = \Theta\!\left( \overline{\mu}_k\!\left( t^{(k)} \right) \right), \quad t^{(k)} = 0, 1, \ldots .
\label{eq:B_k}
\end{equation}
Just as an example, we point out that  an eligible choice for the $\Theta(\cdot)$ function in Eq. \eqref{eq:B_k} is the following one:
\begin{equation}
\Theta\!\left( \overline{\mu}_k\!\left( t^{(k)} \right) \right) = B_0 \times \left( \overline{\mu}_k\!\left( t^{(k)} \right) \right)^{\gamma},
\label{eq:Theta_example}
\end{equation}
where $B_0$ is a positive threshold reference value and $\gamma$ is a non-negative power-shaping coefficient.

\paragraph{Distributed detection of communication failures and time-stamp management}

\textit{Assumption 2} points out that, in the reference scenario of Fig.\ \ref{fig:scenario}, packet-loss events are random and (typically) unpredictable. Hence, in order to detect them in a distributed way, $coworker\#k$ starts its local $Timer_k$ whenever it sends a new model update to the central server (see $line\#9$ of Algorithm \ref{alg:AFAFed_protocol}). By doing so, $coworker\#k$ autonomously detects a packet-loss event whenever $Timer_k$ expires and feedback has been not still received from the central server (see $line\#11$ of Algorithm \ref{alg:AFAFed_protocol}). Interestingly enough, the re-transmission policy implemented by \textit{AFAFed} is of \textit{non}-persistent type \cite{kumar2004}. This means that, after detecting a packet-loss, $coworker\#k$ \textit{refrains} to immediately attempt a retransmission of the loss packet, but it starts a new cluster of local iterations (see the {\sffamily\bfseries go to} statement in $line\#12$ of Algorithm \ref{alg:AFAFed_protocol}). This design choice aims to randomize the transmission times of different coworkers, in order to avoid multiple consecutive packet-loss events in communication-congested FL scenarios.

Finally, we anticipate that, in order to allow the central server to perform model aggregation in an age-aware way (see Eq. (44)), $coworker\#k$ manages a local $timestamp_k$ variable for storing the value of the global time index $t$ at which $coworker\#k$ received the last global model $\vv{\overline{w}}^{(LAST)}$. By doing so, after receiving a new global model from the central server, $coworker\#k$: (i) updates $\vv{\overline{w}}^{(LAST)}$ and $timestamp_k$ (see $line\#31$ of Algorithm \ref{alg:AFAFed_protocol}); (ii) sets its local model $\vv{w}_k$ to $\vv{\overline{w}}^{(LAST)}$ (see $line\#32$ of Algorithm \ref{alg:AFAFed_protocol}); and, then: (iii) starts a new cluster of local primal/dual iterations (see $line\#33$ of Algorithm \ref{alg:AFAFed_protocol}). After finishing the current iteration cluster, $coworker\#k$ communicates the last updated 3-ple: $\left\{ \vv{w}_k, \, \overline{\mu}_k, \, timestamp_k \right\}$ to the central server (see $line\#9$ of Algorithm \ref{alg:AFAFed_protocol}).

Overall, the resulting per-iteration computational effort to be sustained by $coworker\#k$ scales up as  $\mathcal{O}(l+6)$, where $l$ is the size of the vector update in Eq. \eqref{eq:SGD_iteration_1}, while the additive term accounts for the scalar updates in Eqs. \eqref{eq:SGD_iteration_2}, \eqref{eq:mu_average}, \eqref{eq:Omega_max_log}, \eqref{eq:eta_update_1}, \eqref{eq:eta_update_2}, and \eqref{eq:Theta_example}.

\subsection{AFAFed procedures at the central server}
\label{sec:AFAFed_procedures_CS}

The main task of the central server is to adaptively tune the set of fairness coefficients $\left\{ \lambda_k \right\}$ (see $lines\#18-24$ of Algorithm \ref{alg:AFAFed_protocol}), as well as the mixing parameter $\beta$ and the global model $\vv{\overline{w}}$ (see $lines\#25-26$ of Algorithm \ref{alg:AFAFed_protocol}). In order to perform this task in an age-aware way, let:
\begin{equation}
ac(t + 1): \left\{ 1, 2, \ldots, T \right\} \longrightarrow \left\{ 1, \ldots, K \right\},
\label{eq:}
\end{equation}
be an auxiliary function, which is implemented at the central server and returns the index of the coworker from which the central server receives the model update at the global time index $(t+1)$, $t = 0, 1, \ldots, T$\footnote{From a formal point of view, we use $(t+1)$ in place of $t$, because, in the \textit{AFAFed} framework, the central server is not updated at the bootstrapping instant $t = 0$.}.

\paragraph{Adaptive distributed tuning of fairness coefficients}

Let us assume that, at the global time index $t+1$, the central server receives a new updated 3-ple: $\left\{ \vv{w}_{ac(t+1)}, \, \overline{\mu}_{ac(t+1)}, \, timestamp_{ac(t+1)} \right\}$ from the $ac(t+1)$-th coworker. The pursued principle behind the proposed adaptive design of the \textit{AFAFed} fairness coefficients is to scale up (resp., scale down) the current value of $\lambda_{ac(t+1)}$ when the corresponding value $\overline{\mu}_{ac(t+1)}$ of the averaged Lagrange multiplier in Eq. \eqref{eq:mu_average} is larger (resp., smaller) than a suitable upper (resp., lower) fairness threshold. The rationale is that too high (resp., too small) values of $\overline{\mu}_{ac(t+1)}$ indicate that the $ac(t+1)$-th coworker is receiving a too unfair (resp., too biased) service from the central server, i.e. the current global model $\vv{\overline{w}}(t)$ is too far from (resp., too close to) the current local model $\vv{w}_{ac(t+1)}$ of the $ac(t+1)$-th coworker. However, due to both dynamically time-varying stream nature of new arrivals and coworker heterogeneity, the aforementioned upper and lower fairness thresholds should retain the following two main properties. First, at each $t$, they should measure the \textit{average} level of fairness globally experienced by the \textit{full} set of coworkers. Second, these thresholds should be \textit{dynamically} tuned, in order to track the current fairness trend of the coworkers. 

In order to attain the first property, let us assume that the central server implements and updates the following global ensemble average:
\begin{equation}
\widetilde{\mu}(t + 1) \triangleq \frac{1}{t + 1} \sum_{\tau = 1}^{t+1} \overline{\mu}_{ac(\tau)}, \quad t = 0, 1, \ldots,
\label{eq:mu_tilde}
\end{equation}
and the corresponding global ensemble standard deviation:
\begin{equation}
\sigma_{\tilde{\mu}}(t + 1) \triangleq \frac{1}{t + 1} \sum_{\tau = 1}^{t+1} \left| \, \widetilde{\mu}(\tau) - \overline{\mu}_{ac(\tau)} \right|, \quad t = 0, 1, \ldots,
\label{eq:sigma_tilde}
\end{equation}
of the received sequence of per-coworker average Lagrange multipliers in \eqref{eq:mu_average}. Hence, we \textit{propose} that, at each $t$, the central server adaptively tunes the upper: $TH_U(t)$ and lower: $TH_L(t)$ fairness thresholds as follows (see $line\#18$ of Algorithm \ref{alg:AFAFed_protocol}):
\begin{equation}
TH_U(t) \equiv \left| \, \widetilde{\mu}(t) + 4.0 \, \sigma_{\tilde{\mu}}(t) \right|,
\label{eq:TH_U}
\end{equation}
and
\begin{equation}
TH_L(t) \equiv \left| \, \widetilde{\mu}(t) - 4.0 \, \sigma_{\tilde{\mu}}(t) \right|,
\label{eq:TH_L}
\end{equation}
where the terms: $\pm 4.0 \, \sigma_{\tilde{\mu}}(t)$ act as safety margins. Afterwards, as reported in $lines\#20-24$ of Algorithm \ref{alg:AFAFed_protocol}, the central server scales up the current value of $\lambda_{ac(t+1)}$ as in:
\begin{equation}
\lambda_{ac(t+1)} \times \Psi\!\left( \overline{\mu}_{ac(t+1)}, \widetilde{\mu}(t) \right), \quad \text{if } ~ \overline{\mu}_{ac(t+1)} > TH_U(t), 
\label{eq:lambda_scale_up}
\end{equation}
or scale down it as in: 
\begin{equation}
\lambda_{ac(t+1)} \times V\!\left( \overline{\mu}_{ac(t+1)}, \widetilde{\mu}(t) \right), \quad \text{if } ~ \overline{\mu}_{ac(t+1)} < TH_L(t), 
\label{eq:lambda_scale_down}
\end{equation}
where:
\begin{equation}
\Psi\!\left( \overline{\mu}, \widetilde{\mu} \right): \mathbb{R}_0^+ \times \mathbb{R}_0^+ \rightarrow [0, \, +\infty), \quad \text{and} \quad V\!\left( \overline{\mu}, \widetilde{\mu} \right): \mathbb{R}_0^+ \times \mathbb{R}_0^+ \rightarrow (0, \, 1],
\label{eq:Psi_V}
\end{equation}
are two scaling functions, which meet the following defining properties:
\begin{enumerate}
	\item $\Psi\!\left( \overline{\mu}, \widetilde{\mu} \right)$ assumes values in $[1, \, +\infty)$ (i.e., it is a scaling-\textit{up} function). It does not decreases for increasing values of $\left( \overline{\mu} - \widetilde{\mu} \right) \geq 0$; while,
	\item $V\!\left( \overline{\mu}, \widetilde{\mu} \right)$ assumes values in $(0, \, 1]$ (i.e., it is a scaling-\textit{down} function). It does not increase for increasing values of $\left( \overline{\mu} - \widetilde{\mu} \right) \geq 0$.
\end{enumerate}
Examples of eligible up/down scaling functions are the following ones:
\begin{equation}
\Psi\!\left( \overline{\mu}, \widetilde{\mu} \right) \equiv \frac{1}{V\!\left( \overline{\mu}, \widetilde{\mu} \right)} = 1 + \log\!\left( 1 + \frac{\left| \overline{\mu} - \widetilde{\mu} \right|}{1 + \widetilde{\mu}} \right).
\label{eq:example_up_down_func}
\end{equation}
Therefore, after performing the scaling in Eqs. \eqref{eq:lambda_scale_up} and \eqref{eq:lambda_scale_down}, the central server re-normalizes the full set $\Big\{ \lambda_j, \, j = 1, \ldots, K \Big\}$ of the fairness coefficients to be unit-sum as in (see $line\#21$ and $line\#23$ of Algorithm \ref{alg:AFAFed_protocol}):
\begin{equation}
\lambda_j^{(NEW)} = \frac{\lambda_j}{\sum_{m=1}^K \lambda_m}, \quad j = 1, \ldots, K.
\label{eq:lambda_new}
\end{equation}
Finally, the server broadcasts the new set of fairness coefficients $\left\{ \lambda_j^{(NEW)} \right\}$ to all coworkers (see $line\#27$ of Algorithm \ref{alg:AFAFed_protocol}). So doing, it is guaranteed that central server and coworkers share the same updated set of $\lambda$'s coefficients (see $line\#36$ of Algorithm \ref{alg:AFAFed_protocol}).

\paragraph{Age and fairness-aware tuning of the model-mixing parameter}

A \textit{new} feature of the pursued criterion for the design of the updating relationship of the mixing parameter $\beta(t)$ in Eqs. \eqref{eq:CS_update_2_1} and \eqref{eq:CS_update_2_2} is that it \textit{jointly} accounts for the model age and fairness coefficients, according to the following three considerations. First, $\beta(t)$ should scale up/down proportionally the current value $\lambda_{ac(t+1)}$ of the fairness coefficient of the communicating coworker, in order to enforce fairness in the corresponding updating of the global model in Eq. \eqref{eq:CS_update_2_1}. Second, under stationary (resp., non-stationary) application environments, $\beta(t)$ should asymptotically vanish (resp., approach a non-zero positive value), in order to guarantee a stable behavior in the steady-state (resp., quick responsiveness to abrupt environmental changes). Third, since too aged (i.e., stale) model updates may be less reliable, $\beta(t)$ should scale down for increasing values of the \textit{age}  of the current model update, which is formally defined as follows \cite{Zhang2015a,Zhang2016}:
\begin{equation}
age_{ac(t + 1)} \triangleq (t + 1) - timestamp_{ac(t+1)} - 1.
\label{eq:age}
\end{equation}

In order to formalize the above design criteria, let:
\begin{equation}
\Phi(age): \mathbb{N}_0 \longrightarrow \left[ 0, \, 1 \right],
\label{eq:Phi_age}
\end{equation}
be a non-negative function that does not decrease for increasing values of $age$, like, for example, the following ones \cite{Xie2020}: 
\begin{equation} 
\Phi(age) \equiv \left( 1 + age \right)^{-\alpha}, ~ \alpha > 0; \quad \Phi(age) \equiv e^{- \alpha \times age}, ~ \alpha > 0, \quad \text{and:} \quad \Phi(age) \equiv \begin{cases}
1, & 0 \leq age \leq b, \\[1ex]
(1 + age)^{- \alpha}, & age \geq b.
\end{cases}
\label{eq:Phi_age_example}
\end{equation}
Hence, we \textit{propose} the above relationship for the adaptive age and fairness-aware updating of $\beta(\cdot)$ in $line\#25$ of Algorithm \ref{alg:AFAFed_protocol}: 
\begin{equation}
\beta(t) = \max\left\{ \beta_{MIN}; \, \min\left\{ \beta_{MAX}; \, \frac{\lambda_{ac(t+1)} \times \Phi\!\left( age_{ac(t+1)} \right)}{\left( 1 + t \right)^{de}} \right\} \right\}, \quad t = 0, 1, \ldots,
\label{eq:beta_t}
\end{equation}
where: (i) $de$ is a non-negative decay-exponent\footnote{Vanishing (resp., non-vanishing) values of the $de$ exponent may be suitable under stationary (resp., nonstationary) operating environments.}; while, (ii) $\beta_{MIN}$ and $\beta_{MAX}$ with $0 < \beta_{MIN} \leq \beta_{MAX} \leq 1$ are lower/upper clipping parameters.

After updating $\beta(\cdot)$, the central server: (i) performs model aggregation as in Eqs. \eqref{eq:CS_update_2_1} and \eqref{eq:CS_update_2_2} (see $line\#26$ of Algorithm \ref{alg:AFAFed_protocol}); (ii) sends a timestamped copy (with time stamp $(t+1)$) of the updated global model $\vv{\overline{w}}(t+1)$ to the currently communicating $ac(t+1)$-th coworker; (iii) increases its time index $t$; and, finally, (iv) stalls for waiting for a new model update (see $line\#28$ of Algorithm \ref{alg:AFAFed_protocol}).

Overall, the resulting per-aggregation computational effort to be sustained by the central server scales up as $\mathcal{O}(l+4)$, where $l$ is the size of the vector updating in Eq. \eqref{eq:CS_update_2_1}, while the additive term accounts for the scalar updates in Eqs. \eqref{eq:eta_update_1}, \eqref{eq:lambda_scale_up}, \eqref{eq:lambda_scale_down}, \eqref{eq:lambda_new}, and \eqref{eq:beta_t}.

\section{\textit{AFAFed} Convergence Analysis}
\label{sec:AFAFed_convergence}

The goal of this section is to prove the convergence (in expectation) of the sequence $\left\{ \vv{\overline{w}}(t), \, t \geq 0 \right\}$ of the global models evaluated by the central server. In this regard, in order to account for the packet-loss events affecting the uplinks of Fig.\ \ref{fig:scenario}, we indicate by $P_k$, $k = 1, \ldots, K$, the steady-state probability of successful transmissions from $coworker\#k$, formally defined as follows (see Eq. \eqref{eq:csi_k}): 
\begin{equation}
P_k \triangleq E\!\left\{ \xi_k(t) \right\} \equiv \lim_{t \rightarrow \infty} \frac{1}{t} \sum_{\tau=1}^t \xi_k(\tau), \quad k = 1, \ldots, K.
\label{eq:P_k}
\end{equation}
Since $P_k$ vanishes when the corresponding packet loss probability in Eq. \eqref{eq:P_loss} approaches the unit (see Eq. \eqref{eq:csi_k}), we refer to the set $\left\{ P_k \right\}$ as the coworker access probabilities.

Regarding the methodology pursued for proving the convergence of the sequence $\left\{ \vv{\overline{w}}(t) \right\}$, let us consider, as a benchmark, the \textit{Sy}nchronous \textit{C}entralized \textit{G}radient-\textit{D}escent scheme (\textit{SyCGD}), in which: (i) \textit{all} the training data $S \triangleq \cup_{k=1}^K S_k$ stays at the central server; and, (ii) the \textit{full} gradient in Eq. \eqref{eq:DF} is used to carry out the following \textit{synchronous} gradient-descent iterations: 
\begin{equation}
\vv{\overline{w}}(t+1) = \vv{\overline{w}}(t) - \beta(t) \vv{\nabla}F\!\left( \vv{\overline{w}}(t) \right), \quad t = 0, 1, \ldots
\label{eq:sync_iteration}
\end{equation}
Hence, a comparison of the asynchronous SGD iterations of Eq.\eqref{eq:CS_update_2_2} and the \textit{SyCGD} ones of Eq. \eqref{eq:sync_iteration} supports the conclusion that asynchronous and synchronous iterations may follow similar time-paths, provided that the stochastic gradient $\widehat{\vv{G}}(\cdot)$ in Eq. \eqref{eq:G} is a reliable estimate of the corresponding full gradient $\vv{\nabla}F(\cdot)$ in Eq. \eqref{eq:sync_iteration}. Hence, we expect that, if we would capable to suitably limit the gap between these two gradients, then, the convergence rate of the asynchronous SGD iterations in Eqs. \eqref{eq:CS_update_2_1} and \eqref{eq:CS_update_2_2} would approach the one of the \textit{SyCGD} benchmark in Eq. \eqref{eq:sync_iteration}.

\subsection{Preliminaries and baseline properties}
\label{sec:baseline_properties}

Motivated by the above expectation, we introduce the following baseline assumptions on the formal properties of the considered local and global loss functions:
\begin{assumption}[\textit{Baseline assumptions on the global/local loss functions}]
By referring to the defining relationships in Eqs. \eqref{eq:F_k}--\eqref{eq:F_global}, we assume that:
\begin{enumerate}
	\item there exists at least a global minimizer in Eq. \eqref{eq:w_star} with finite global minimum Eq. \eqref{eq:F_star} in correspondence of the considered (possibly, \textbf{non-convex}) local functions $\left\{ F_k(\cdot), \, k = 1, \ldots, K \right\}$;
	\item the local loss functions are $\zeta$-smooth, i.e., there exists a positive scalar constant $\zeta$ such that:
	\begin{equation}
	\left\| \vv{\nabla}F_k\!\left( \vv{w}_1 \right) - \vv{\nabla}F_k\!\left( \vv{w}_2 \right) \right\| \leq \zeta \left\| \vv{w}_1 - \vv{w}_2 \right\|, \quad \forall \vv{w}_1, \vv{w}_2 \in \mathbb{R}^l, ~ k = 1, \ldots, K;
	\label{eq:zeta_smooth}
	\end{equation}
	\item the stochastic gradients in Eq. \eqref{eq:DF_tilde_k} provide unbiased estimates of the corresponding full gradients in Eq. \eqref{eq:DF_k}, that is:
	\begin{equation}
	E_{\mathcal{D}_k}\left\{ \vv{\nabla}\widetilde{F}_k\!\left( \vv{w} \right) \right\} \equiv \vv{\nabla}F_k\!\left( \vv{w} \right), \quad \forall \vv{w} \in \mathbb{R}^l, ~ k = 1, \ldots, K;
	\label{eq:E_D_k}
	\end{equation}
	\item the variances of the stochastic gradients are bounded and scale down with the sizes of the corresponding mini-batches as in:
	\begin{equation}
	E_{\mathcal{D}_k}\left\{ \left\| \vv{\nabla}\widetilde{F}_k\!\left( \vv{w} \right) - \vv{\nabla}F_k\!\left( \vv{w} \right) \right\|^2 \right\} \leq \frac{\sigma_{k,1}^2}{\left| MB_k \right|}, \quad \forall \vv{w} \in \mathbb{R}^l, ~ k = 1, \ldots, K,
	\label{eq:E_D_k_var_1}
	\end{equation}
	where $\sigma_{k,1}^2 \geq 0$ is the variance of the noise affecting the $k$-th stochastic gradient when the mini-batch is unit-sized;
	\item under any setting of the fairness coefficients $\left\{ \lambda_j \right\}$, the gradient variances induced by non-i.i.d. data are limited as follows:
	\begin{equation}
	E_{\mathcal{D}_k}\left\{ \left\| \vv{\nabla}F_k\!\left( \vv{w} \right) - \vv{\nabla}F\!\left( \vv{w} \right) \right\|^2 \right\} \leq \sigma_{k,2}^2, \quad \forall \vv{w} \in \mathbb{R}^l, ~ k = 1, \ldots, K.
	\label{eq:E_D_k_var_2}
	\end{equation}
\end{enumerate}
\vspace{-1em}\hfill $\blacksquare$
\end{assumption}
\noindent Although quite standard (see, for example, \cite{Boyd2011,Xie2020,Wang2019}), the above assumptions merit some auxiliary remarks.

First, as formally proved, for example, in \cite{Wang2019}, the $\zeta$-smooth property of the local loss functions in Eq. \eqref{eq:zeta_smooth} implies that also the resulting global function in Eq. \eqref{eq:F_global} is $\zeta$-smooth, so that we have: 
\begin{equation}
\left\| \vv{\nabla}F\!\left( \vv{w}_1 \right) - \vv{\nabla}F\!\left( \vv{w}_2 \right) \right\| \leq \zeta \left\| \vv{w}_1 - \vv{w}_2 \right\|, \quad \forall \vv{w}_1, \vv{w}_2 \in \mathbb{R}^l, ~ \forall \left\{ \lambda_j \right\}.
\label{eq:zeta_smooth_F}
\end{equation}

Second, from the bounds in Eqs. \eqref{eq:E_D_k_var_1} and \eqref{eq:E_D_k_var_2}, it follows that the variances of the local stochastic gradients in Eq. \eqref{eq:DF_tilde_k} around the full global gradient in Eq. \eqref{eq:DF} are limited as in (see the Appendix \ref{appendix:B} for the proof):
\begin{equation}
E_{\mathcal{D}_k}\left\{ \left\| \vv{\nabla}\widetilde{F}_k\!\left( \vv{w} \right) - \vv{\nabla}F\!\left( \vv{w} \right) \right\|^2 \right\} \leq \Sigma_k^2, \quad \forall \vv{w} \in \mathbb{R}^l, ~ k = 1, \ldots, K,
\label{eq:E_D_k_var_3}
\end{equation}
where:
\begin{equation}
\Sigma_k^2 \triangleq \frac{\sigma_{k,1}^2}{\left| MB_k \right|} + \sigma_{k,2}^2, \quad k = 1, \ldots, K,
\label{eq:Sigma_k}
\end{equation}
accounts for both mini batch-induced and data skewness-induced gradient variances. 

Finally, we remark that the introduced \textit{Assumption 3} allows us to define three main ensemble averages over the spectrum of the access probabilities in Eq. \eqref{eq:P_k}. The first index is the ensemble average $\overline{\Sigma^2}$ of the per-coworker variances of Eq. \eqref{eq:Sigma_k}, formally defined as follows:
\begin{equation}
\overline{\Sigma^2} \triangleq E_{\xi \sim P}\left\{ \Sigma_k^2 \right\} \equiv \sum_{k=1}^K P_k \Sigma_k^2.
\label{eq:Sigma_average}
\end{equation}
The second and third indexes are the first $\overline{I}$ and second $\overline{I^2}$ moments of the numbers of per-cluster local iterations performed by the coworkers, i.e.,
\begin{equation}
\overline{I} \triangleq E_{\xi \sim P}\left\{ Iter_k \right\} \equiv \sum_{k=1}^K P_k \overline{Iter_k} , \quad \text{and:} \quad \overline{I^2} \triangleq E_{\xi \sim P}\left\{ Iter_k^2 \right\} \equiv \sum_{k=1}^K P_k \overline{Iter_k^2},
\label{eq:I_average}
\end{equation}
where: $\overline{Iter_k} \triangleq E_{\mathcal{D}_k}\!\left\{ Iter_k \right\}$, and: $\overline{Iter_k^2} \triangleq E_{\mathcal{D}_k}\!\left\{ Iter_k^2 \right\}$ are the corresponding first and second moments of the r.v. $Iter_k$ in Eq. \eqref{eq:Iter_k}.

\subsection{Convergence assumptions and their meaning/roles}
\label{sec:convergence_assumptions}

State-of-the-art literature \cite{Lian2015} has already pointed out that, under non-convex settings, commonly metrics used for evaluating the convergence rate in convex optimization (like: $F\!\left(\vv{\overline{w}}(t) \right) - F^\ast$, or $\left\| \vv{\overline{w}} - \vv{\overline{w}}^\ast \right\|$) are no longer eligible. Hence, according to, for example, \cite{Lian2015,Feyzmahdavian2016,Peng2019}, we resort to the convergence (in expectation) of the sequence: $\left\{ \left\| \vv{\nabla}F\!\left( \vv{\overline{w}} \right) \right\|^2, \, t \geq 0 \right\}$ along the path: $\left\{ \vv{\overline{w}}(t), \, t \geq 0 \right\}$ followed by the  iterations in Eqs. \eqref{eq:CS_update_2_1} and \eqref{eq:CS_update_2_2} as the metric for evaluating the \textit{AFAFed} convergence. This means, in turn, that our target is to formally characterize the convergence rate of the following global average:
\begin{equation}
E\!\left\{ \frac{1}{T} \sum_{t=0}^{T-1} \left\| \vv{\nabla}F\!\left( \vv{\overline{w}}(t) \right) \right\|^2 \right\},
\label{eq:global_average}
\end{equation}
for increasing values of $T$. The pursued formal methodology is inspired by the so called \textit{Gradient Coherence} principle in \cite{Dai2019}. It measures the similitude between the stochastic: $\widehat{\vv{G}}(\cdot)$ and full: $\vv{\nabla}F(\cdot)$ gradients in Eqs. \eqref{eq:CS_update_2_2} and \eqref{eq:sync_iteration} through the corresponding average gap between the cross-product: $\widehat{\vv{G}}(\cdot)^T \, \vv{\nabla}F(\cdot)$ and the corresponding benchmark value: $\vv{\nabla}F(\cdot)^T \, \vv{\nabla}F(\cdot) \equiv \left\| \vv{\nabla}F(\cdot) \right\|^2$. Hence, motivated by this consideration, we introduce the following formal assumption:
\begin{assumption}[\textit{Assumptions on the gradient coherence}]
By referring to the stochastic and full gradient-descent iterations in Eqs. \eqref{eq:CS_update_2_2} and \eqref{eq:sync_iteration}, let us assume that:
\begin{enumerate}
	\item there exist two scalar positive constants $\Gamma \geq C > 0$ such that:
	\begin{equation}
	\left( E_{\xi \sim P}\!\left\{ \widehat{\vv{G}}\!\left( \vv{\overline{w}}; \left\{ \xi_k \right\} \right) \right\} \right)^T  \vv{\nabla}F\left( \vv{\overline{w}} \right) \geq C \left\| \vv{\nabla}F\!\left( \vv{\overline{w}} \right) \right\|^2, \quad \forall \vv{\overline{w}}, ~ \forall \left\{ \lambda_k \right\},
	\label{eq:assump_4_1}
	\end{equation}
	and:
	\begin{equation}
	\left\|  E_{\xi \sim P}\!\left\{ \widehat{\vv{G}}\!\left( \vv{\overline{w}}; \left\{ \xi_k \right\} \right) \right\}  \right\| \leq \Gamma \left\| \vv{\nabla}F\!\left( \vv{\overline{w}} \right) \right\|, \quad \forall \vv{\overline{w}}, ~ \forall \left\{ \lambda_k \right\};
	\label{eq:assump_4_2}
	\end{equation}
	\item there exists a non-negative scalar constant $A \geq 0$ such that the stochastic-gradient variance:
	\begin{equation}
	\sigma_{\widehat{\vv{G}}}^2 \triangleq var\!\left\{ \widehat{\vv{G}} \right\} \equiv E_{\xi \sim P}\!\left\{ \left\| \widehat{\vv{G}}\!\left( \cdot \right) \right\|^2 \right\} - \left\| E_{\xi \sim P}\!\left\{ \widehat{\vv{G}}\!\left( \cdot \right) \right\} \right\|^2,
	\label{eq:assump_4_3}
	\end{equation}
	is upper bounded as follows:
	\begin{equation}
	\sigma_{\widehat{\vv{G}}}^2 \leq A + \left\| \vv{\nabla}F\!\left( \vv{\overline{w}} \right) \right\|^2, \quad ~ \forall \left\{ \lambda_k \right\}.
	\label{eq:assump_4_4}
	\end{equation}
\end{enumerate}
\vspace{-1em}\hfill $\blacksquare$
\end{assumption}
\noindent The meaning/role and feasible values of the constants $\Gamma$, $C$, and $A$ involved in the above \textit{Assumption 4} are featured by the following Lemma \ref{lem:1} (see Appendix \ref{appendix:B} for the proof):
\begin{lemma}[\textit{On the meaning/role and feasible values of the constants $\Gamma$, $C$, and $A$}]\label{lem:1}
Let us assume that the global stochastic gradient $\widehat{\vv{G}}\!\left( \cdot \right)$ in Eq. \eqref{eq:G} provides a (possibly, biased) estimate of the full gradient $\vv{\nabla}F(\cdot)$ in Eq. \eqref{eq:F_global}, i.e., let us assume that: $\widehat{\vv{G}}\!\left( \cdot \right) \equiv \vv{\nabla}F(\cdot) + \vv{n}$, where $\vv{n} \in \mathbb{R}^l$ is a noise r.v., which meets the following two assumptions:
\begin{equation}
E_{\xi \sim P}\!\left\{ \vv{n} \right\} = \gamma_0 \vv{\nabla}F(\cdot),
\label{eq:lemma_1_1}
\end{equation}
with $\gamma_0 > -1$, and,
\begin{equation}
E_{\xi \sim P}\!\left\{ \left\| \vv{n} \right\|^2 \right\} = \gamma_1 \left\| \vv{\nabla}F(\cdot) \right\|^2 + b_0,
\label{eq:lemma_1_2}
\end{equation}
with $b_0 \geq 0$ and $0 \leq \left( \gamma_1 - \gamma_0^2 \right) \leq 1$. Then, the spectra of feasible values of $C$, $\Gamma$, and $A$ in Eqs. \eqref{eq:assump_4_1}--\eqref{eq:assump_4_4} are related to the $\gamma_0$, $\gamma_1$, and $b_0$ parameters as follows: 
\begin{equation}
0 < C \leq \left( 1 + \gamma_0 \right), \quad \Gamma \geq \left( 1 + \gamma_0 \right), \quad \text{and} \quad A \geq b_0.
\label{eq:lemma_1_3}
\end{equation}
Furthermore, when $\widehat{\vv{G}}\!\left( \cdot \right)$ provides an unbiased estimate of $\vv{\nabla}F(\cdot)$, the relationships in Eq. \eqref{eq:lemma_1_3} still hold with $\gamma_0 = \gamma_1 = 0$.
\vspace{-1em}\hfill $\blacksquare$
\end{lemma}

\vspace{2em}
\noindent The bounds reported in Eq. \eqref{eq:lemma_1_3} highlight how the constants $C$, $\Gamma$ and $A$ in \textit{Assumption 4} are, indeed, related to the statistical properties of the noise affecting the stochastic global gradient $\widehat{\vv{G}}\!\left( \cdot \right)$.

\subsection{The \textit{AFAFed} convergence bounds}
\label{sec:convergence_bounds}

In Appendix \ref{appendix:C}, the following main result is proved:
\begin{proposition}[\textit{AFAFed convergence rate}]\label{prop:1}
Let us assume that $\beta_{MAX}$ in Eq. \eqref{eq:beta_t} is limited as in:
\begin{equation}
\beta_{MAX} \leq \frac{2 C \varepsilon}{\zeta \left( 1 + \Gamma^2 \right)},
\label{eq:prop_1_1}
\end{equation}
where $\varepsilon$ is a regularizing parameter, which may be set by the central server over the open interval $(0, \, 1)$. Then, under the (previously reported) Assumptions 1--4, the following upper bound holds on the convergence rate of the \textit{AFAFed} Algorithm \ref{alg:AFAFed_protocol}:
\begin{equation}
E\!\left\{ \frac{1}{T} \sum_{t=0}^{T-1} \beta(t) \left\| \vv{\nabla}F\!\left( \vv{\overline{w}}(t) \right) \right\|^2 \right\} \leq \frac{E\!\left\{ F\!\left( \vv{\overline{w}}(0) \right) \right\} - F^\ast}{C \left( 1 - \varepsilon \right) T} + \frac{A \zeta}{2C \left( 1 - \varepsilon \right)} \, E\!\left\{ \frac{1}{T} \sum_{t=0}^{T-1} \beta(t)^2 \right\}.
\label{eq:prop_1_2}
\end{equation}
Furthermore, in the case in which $\beta_{MIN}$ in Eq. \eqref{eq:beta_t} is positive, we have that:
\begin{equation}
E\!\left\{ \frac{1}{T} \sum_{t=0}^{T-1} \left\| \vv{\nabla}F\!\left( \vv{\overline{w}}(t) \right) \right\|^2 \right\} \leq \frac{E\!\left\{ F\!\left( \vv{\overline{w}}(0) \right) \right\} - F^\ast}{C \left( 1 - \varepsilon \right) \beta_{MIN} T} + \frac{A \zeta \beta_{MAX}^2}{2C \left( 1 - \varepsilon \right) \beta_{MIN}},
\label{eq:prop_1_3}
\end{equation}
which, in turn, in the case of constant $\beta$ (i.e., in the case of $\beta_{MAX} \equiv \beta_{MIN}$) reduces to:
\begin{equation}
E\!\left\{ \frac{1}{T} \sum_{t=0}^{T-1} \left\| \vv{\nabla}F\!\left( \vv{\overline{w}}(t) \right) \right\|^2 \right\} \leq \frac{E\!\left\{ F\!\left( \vv{\overline{w}}(0) \right) \right\} - F^\ast}{C \left( 1 - \varepsilon \right) \beta T} + \frac{A \zeta \beta}{2C \left( 1 - \varepsilon \right)}.
\label{eq:prop_1_4}
\end{equation}
\vspace{-1em}\hfill $\blacksquare$
\end{proposition}

\vspace{2em}
The reported bounds merit two main explicative remarks.
First, all the bounds in Eqs. \eqref{eq:prop_1_2}--\eqref{eq:prop_1_4} are the summation of two non-negative terms. The first terms guarantee that the convergence rate of Algorithm \ref{alg:AFAFed_protocol} is (at least): $\mathcal{O}\left(1/T\right)$ for increasing $T$, while the second terms at the r.h.s.'s of Eqs. \eqref{eq:prop_1_2}--\eqref{eq:prop_1_4} fix the \textit{AFAFed} convergence performance for $T \rightarrow \infty$. They point out that the asymptotic convergence is guaranteed up to a spherical neighborhood of a stationary point of the global loss function $F(\cdot)$, whose squared radius scales up (resp., down) with the product $A \zeta$ (resp., $C \left( 1 - \varepsilon \right)$). Second, the version of \textit{AFAFed} reported in Algorithm \ref{alg:AFAFed_protocol} refers to local primal/dual iterates that are implemented through first-order SGD. However, the analytical derivation in Appendices \ref{appendix:B} and \ref{appendix:C} point out that the reported convergence bounds still \textit{apply verbatim} under any arbitrary local solver, \textit{provided} that the corresponding $C$, $\Gamma$, and $A$ parameters of \textit{Assumption 4} refers to the actually implemented solver.

\subsection{Scaled forms of the convergence bounds}
\label{sec:scaling_bounds}

Our next step is to evaluate the impact of the ensemble-averaged global indexes in Eqs. \eqref{eq:Sigma_average} and \eqref{eq:I_average} and the average packet-loss probability $\overline{P}^{\,(LOSS)}$ of the uplinks of Fig.\ \ref{fig:scenario} on the convergence bounds of Eqs. \eqref{eq:prop_1_2}--\eqref{eq:prop_1_4}. To this end, let us indicate by $C_0$, $\Gamma_0$, and $A_0$ the values assumed by the parameters $C$, $\Gamma$ and $A$ in Eqs. \eqref{eq:assump_4_1}--\eqref{eq:assump_4_4} under the following normalized operating conditions: $\overline{\Sigma^2} = \overline{I} = \overline{I^2} = 1$. Therefore, the developments reported in \ref{appendix:B} lead to the conclusion that $C$, $\Gamma$, and $A$ scale with $\overline{\Sigma^2}$, $\overline{I}$, and $\overline{I^2}$ as in:
%
\begin{subequations}\label{eq:bound_scale}
  \allowdisplaybreaks
	\begin{align}
	C &= C_0 \overline{I},  \label{eq:bound_scale_1} \\[1ex]
	\Gamma &= \Gamma_0 \overline{I},  \label{eq:bound_scale_2} \\
	\intertext{and,}
	A &= A_0 \, \overline{\Sigma^2} \: \overline{I^2}.  \label{eq:bound_scale_3}
	\end{align}
\end{subequations}
%
Hence, by inserting the above relationships in Eqs. \eqref{eq:prop_1_2}--\eqref{eq:prop_1_4}, we formally characterize the dependence of the \textit{AFAFed} convergence bounds on the ensemble-averaged global indexes $\overline{\Sigma^2}$, $\overline{I}$, and $\overline{I^2}$. Just as an example, we may rewrite the bound of Eq. \eqref{eq:prop_1_3} in the following equivalent form: 
\begin{equation}
E\!\left\{ \frac{1}{T} \sum_{t=0}^{T-1} \left\| \vv{\nabla}F\!\left( \vv{\overline{w}}(t) \right) \right\|^2 \right\} \leq \frac{E\!\left\{ F\!\left( \vv{\overline{w}}(0) \right) \right\} - F^\ast}{C_0 \overline{I} \left( 1 - \varepsilon \right) \beta_{MIN} T} + \frac{A \overline{\Sigma^2} \: \overline{I^2} \, \zeta \, \beta_{MAX}^2}{2C_0 \overline{I} \left( 1 - \varepsilon \right) \beta_{MIN}}, \quad \text{with:} \quad \beta_{MAX} \leq \frac{2C_0 \overline{I} \, \varepsilon}{\zeta \left( 1 + \left( \Gamma_0 \overline{I} \right)^2 \right)}.
\label{eq:equivalent_bound}
\end{equation}

In practical application scenarios, the online profiling of the (possibly, \textit{a priori} unknown) parameters $C$, $\Gamma$ and $A$ may be carried out as detailed in the next Section \ref{sec:AFAFed_implementation_aspects}.

\begin{remark}{On the contrasting effects of $\overline{I}$ and $\overline{I^2}$}\label{remark:6}
An application of the Jensen's inequality \cite{kumar2004} guarantees that: $\overline{I^2}/\overline{I} \geq \overline{I}$, so that the second terms at the r.h.s. of the bound in Eq. \eqref{eq:equivalent_bound} scale up (at least) as $\mathcal{O}\left( \overline{I} \right)$ for increasing values of $\overline{I}$. Since the first term at the r.h.s. of this bound scale down as $\mathcal{O}\left( 1/\overline{I} \right)$ for growing $\overline{I}$, we conclude that the effects of $\overline{I}$ on the convergence properties of \textit{AFAFed} are, indeed, contrasting. Specifically, large (resp., small) values of $\overline{I}$ speed up (resp., slow down) the resulting \textit{AFAFed} convergence rate at small $T$'s, but, at the same time, they increase (resp., decrease) the residual average squared fluctuations around the reached stationary-point of the global loss function $F(\cdot)$ for large $T$'s. This supports the conclusion that, since the optimized setting of $\overline{I}$ is typically \textit{a priori} unknown, an adaptive and tuning of the per-cluster number of local iterations carried out by each coworker is, indeed, mandatory under the FL scenario of Fig.\ \ref{fig:scenario}. This provides formal motivation of the choice  pursued in Section \ref{sec:AFAFed_procedures} for the adaptive design of the \textit{AFAFed} in Algorithm \ref{alg:AFAFed_protocol} (see, in particular, Eq. \eqref{eq:Iter_k} and related text). 
\end{remark}

\begin{remark}{On the contrasting effects of the ensemble-averaged variance $\overline{\Sigma^2}$}\label{remark:7}
By definition, $\overline{\Sigma^2}$ in Eq. \eqref{eq:Sigma_average} measures the ensemble-averaged squared jitter of the local gradients in Eq. \eqref{eq:DF_tilde_k} around the global full gradient of Eq. \eqref{eq:DF}, which is caused by both stochastic gradients and data skewness (see Eqs. \eqref{eq:Sigma_k} and \eqref{eq:Sigma_average}). Accordingly, the impact of large values of $\overline{\Sigma^2}$ on the bound in Eq. \eqref{eq:equivalent_bound} is negative, because large $\overline{\Sigma^2}$'s increase the average squared residual errors experienced by \textit{AFAFed} for large $T$'s. However, this (negative) conclusion is somewhat counterbalanced by the analysis carried out in \cite{Jin2017}. Specifically, the main conclusion of this analysis is that, when the Hessian matrix of the global loss function $F(\cdot)$ retains at least one negative eigenvalue at each saddle point $\overline{\vv{w}}_{sdl}$, then, a random perturbation of $\overline{\vv{w}}_{sdl}$ allows $F(\cdot)$ to escape it with high probability through a finite number of perturbed iterations. Hence, since, in the \textit{AFAFed} framework, the global sequence $\left\{ \vv{\overline{w}}(t) \right\}$ in Eqs. \eqref{eq:CS_update_2_1} and \eqref{eq:CS_update_2_2} is driven by the global stochastic gradient $\widehat{\vv{G}}(\cdot)$ and $var\left\{ \widehat{\vv{G}} \right\}$ in \eqref{eq:assump_4_3} increases as $\mathcal{O}\left( \overline{\Sigma^2} \right)$ (see Eqs. \eqref{eq:assump_4_4} and \eqref{eq:bound_scale_3}), we expect that large enough values of $\overline{\Sigma^2}$ may help the gradient sequence $\left\{ \vv{\nabla}F\!\left( \vv{\overline{w}}(t) \right) \right\}$ to escape saddle points, so to converge to neighborhoods of local/global minima of $F(\cdot)$.
\end{remark}

A last remark concerns the effect of the packet-loss probabilities in Eq. \eqref{eq:P_loss} on the convergence rate of the bound in Eq. \eqref{eq:equivalent_bound}. After noting that the number $T$ of updates performed by the central server does \textit{not} depend, \textit{by definition}, on the probability spectrum $\left\{ P_k^{\,(LOSS)} \right\}$, let $\overline{P}^{(LOSS)}$ be the average packet-loss probability, and let $N^{(TR)(TOT)}$ indicate the total number of successful-plus-failed transmissions performed by all coworkers of Fig.\ \ref{fig:scenario}. Hence, since $T$ in Eq. \eqref{eq:equivalent_bound} equates, by definition, the (average) fraction of the number of the total transmissions which do not undergo packet-loss events, we obtain the following scaling law \cite{kumar2004}:
\begin{equation}
T \propto \left( 1 - \overline{P}^{\,(LOSS)} \right) N^{(TR)(TOT)}.
\label{eq:T_transmission}
\end{equation}
Thus, after inserting Eq. \eqref{eq:T_transmission} at the denominator of the first term of Eq. \eqref{eq:equivalent_bound}, we conclude that the effect of the packet-loss probabilities is to scale down the average convergence rate of the bound by a factor: $\left( 1 - \overline{P}^{\,(LOSS)} \right)$, while they do not modify the attained asymptotic value.

\section{\textit{AFAFed} Implementation Aspects}
\label{sec:AFAFed_implementation_aspects}

The goal of this section is to consider two main implementation aspects of the \textit{AFAFed} Algorithm \ref{alg:AFAFed_protocol}, which concern: (i) the online profiling of the (possibly, \emph{a priori} unknown) parameters involved by the bounds of \textit{Proposition \ref{prop:1}}; and, (ii) the optimized design and implementation of a throughput-maximizing distributed policy for the dynamic management of the streams of arrivals in Fig. \ref{fig:scenario}.

\subsection{Profiling the \textit{AFAFed} parameters}
\label{sec:AFAFed_parameters}

The developed procedure for the online profiling of the parameters involved in the bounds of \textit{Proposition \ref{prop:1}} relies on the following formal results on the feasibility of their estimates (see the last part of Appendix \ref{appendix:B} for the proof): 
\begin{lemma}[Feasible estimates of the \textit{AFAFed} parameters]\label{lem:2}
Let $\overline{\vv{\widehat{G}}}$, $\overline{\left\|\vv{\widehat{G}}\right\|}$, $\overline{\left\|\vv{\widehat{G}}\right\|^2}$, and $\widehat{\vv{\nabla}F}$ indicate estimates of the corresponding parameters: $E_{\xi \sim P}\left\{ \vv{\overline{G}} \right\}$, $\left\| E_{\xi \sim P}\left\{ \vv{\overline{G}} \right\} \right\|$, $E_{\xi \sim P}\left\{ \left\| \vv{\overline{G}} \right\|^2 \right\}$, and $\vv{\nabla}F$. Hence, if and only if the following three conditions are simultaneously met:
%
\begin{subequations}\label{eq:lemma_2_I}
  \allowdisplaybreaks
	\begin{align}
	&\overline{\vv{\widehat{G}}}^T \, \widehat{\vv{\nabla}F} > 0,  \label{eq:lemma_2_1} \\[1ex]
	&k_0 \triangleq \frac{\left\| \overline{\vv{\widehat{G}}} \right\| \, \left\| \widehat{\vv{\nabla}F} \right\|}{\left(\overline{\vv{\widehat{G}}}\right)^T \, \left(\widehat{\vv{\nabla}F}\right)} -1 \geq 0,  \label{eq:lemma_2_2} \\[1ex]
	&\overline{\left\|\vv{\widehat{G}}\right\|^2} - \left( \overline{\left\|\vv{\widehat{G}}\right\|} \right)^2 - \left\| \widehat{\vv{\nabla}F} \right\|^2  \geq 0,  \label{eq:lemma_2_3} 
	\end{align}
\end{subequations}
%
then, the following three relationships:
%
\begin{subequations}\label{eq:lemma_2_II}
  \allowdisplaybreaks
	\begin{align}
	\widehat{C} &\triangleq \frac{\overline{\vv{\widehat{G}}}^T \, \widehat{\vv{\nabla}F}}{\left\| \widehat{\vv{\nabla}F} \right\|^2},  \label{eq:lemma_2_4} \\[1ex]
	\widehat{\Gamma} &\triangleq \left( 1 + k_0 \right) \widehat{C},  \label{eq:lemma_2_5} \\[1ex]
	\widehat{A} &\triangleq \overline{\left\|\vv{\widehat{G}}\right\|^2} - \left( \overline{\left\|\vv{\widehat{G}}\right\|} \right)^2 - \left\| \widehat{\vv{\nabla}F} \right\|^2,  \label{eq:lemma_2_6} 
	\end{align}
\end{subequations}
%
provide feasible estimates of the parameters $C$, $\Gamma$ and $A$ in Eqs. \eqref{eq:assump_4_1}--\eqref{eq:assump_4_4}.
\vspace{-1em}\hfill $\blacksquare$
\end{lemma}

\vspace{2em}
On the basis of Lemma \ref{lem:2}, we conclude that when \textit{a priori} information about the actual parameters $C$, $\Gamma$, $A$ is not available, the only viable means of computing their estimates in Eqs. \eqref{eq:lemma_2_4}--\eqref{eq:lemma_2_6} is to run a (suitably modified version) of the \textit{AFAFed} protocol, in order to perform the online profiling of the estimates: $\overline{\vv{\widehat{G}}}$, $\overline{\left\|\vv{\widehat{G}}\right\|}$, $\overline{\left\|\vv{\widehat{G}}\right\|^2}$, and $\widehat{\vv{\nabla}F}$ on the r.h.s of Eqs. \eqref{eq:lemma_2_4}--\eqref{eq:lemma_2_6}.

\begin{algorithm*}[htbp]
\caption{--- Online profiling of the \textit{AFAFed} parameters}
\label{alg:AFAFed_profiling}
\footnotesize
\begin{algorithmic}[1]
\LeftComment{\textit{\textbf{Skeleton of the modified AFAFed protocol for online parameter profiling}}}
\State {Central server broadcasts the initialization model $\vv{\overline{w}}(0)$ to all coworkers};
\For{$ t = 0 : \left( T - 1 \right) $}
\State {central server and coworkers perform the $t$-th iteration of Algorithm \ref{alg:AFAFed_protocol}};
\If {$coworker\#k$ communicates to the central server}
\State {$coworker\#k$ sends its last computed local stochastic gradient $\vv{\nabla}\widetilde{F}_k\!\left( \vv{w}_k(t+1) \right)$ to the central server};
\State {central server stores the received gradient};
\EndIf \Comment{End-if at Step\#4}
\State {central server evaluates and stores: $\vv{\widehat{G}}(t) = \sum_{k=1}^K \xi_k(t+1) \left[ \vv{\overline{w}}(t) - \vv{w}_k(t+1) \right]$};
\EndFor \Comment{End for at Step\#2}
\Statex {}
\LeftComment{\textit{\textbf{AFAFed profiling phase}}}
\LeftComment{After ending to run Algorithm \ref{alg:AFAFed_protocol}, the central server and coworkers synchronously perform the following steps:}
\State {Central server broadcasts the final global model $\vv{\overline{w}}(T)$};
\ParFor {$ k = 1:K $}
\State {$coworker\#k$ evaluates: $F_k(0) \triangleq F_k\!\left( \vv{\overline{w}}(0) \right)$, $F_k(T) \triangleq F_k\!\left( \vv{\overline{w}}(T) \right)$, and $\widehat{\zeta}_k \triangleq \frac{\left\| \vv{\nabla}\widetilde{F}_k\!\left( \vv{w}_k(T) \right) - \vv{\nabla}\widetilde{F}_k\!\left( \vv{w}_k(0) \right) \right\|}{\left\| \vv{\overline{w}}(T) - \vv{\overline{w}}(0) \right\|}$};
\State {$coworker\#k$ sends the 3-ple: $\left\{ F_k(0), \, F_k(T), \, \widetilde{\zeta}_k \right\}$ of the locally estimated parameters to the central server};
\EndParFor \Comment{End parfor at Step\#11}
\State {Central server evaluates the following four sample averages: 
\begin{flushleft}
\[
\overline{\vv{\widehat{G}}} = \frac{1}{T} \sum_{t=0}^{T-1} \vv{\widehat{G}}(t),
\]
\[
\overline{\left\| \vv{\widehat{G}} \right\|} = \frac{1}{T} \sum_{t=0}^{T-1} \left\| \vv{\widehat{G}}(t) \right\|,
\]
\[
\overline{\left\| \vv{\widehat{G}} \right\|^2}  = \frac{1}{T} \sum_{t=0}^{T-1} \left\| \vv{\widehat{G}}(t) \right\|^2,
\]
and
\[
\widehat{\vv{\nabla}F} = \frac{1}{T} \sum_{k=1}^K \xi_k(t+1) \vv{\nabla}\widetilde{F}_k\!\left( \vv{w}_k(t+1) \right);
\]
\end{flushleft}
}
\State {Central server computes the following estimates: 
\begin{flushleft}
\[
\widehat{F}^\ast \triangleq \sum_{k=1}^K \lambda_k(T) F_k(T);
\]
\[
\widehat{F}(0) \triangleq \sum_{k=1}^K \lambda_k(T) F_k(0);
\]
and
\[
\widehat{\zeta} \triangleq \max_{1 \leq k \leq K} \left\{ \widehat{\zeta}_k \right\},
\]
\end{flushleft}
of the actual parameters: $F^\ast$, $E\!\left\{ F\!\left( \vv{\overline{w}}(0) \right) \right\}$, and $\zeta$, respectively};
\If {conditions in Eqs.\eqref{eq:lemma_2_1}--\eqref{eq:lemma_2_3} are met}
\State {central server computes the estimates: $\widehat{C}$, $\widehat{\Gamma}$, and $\widehat{A}$ as in Eqs. \eqref{eq:lemma_2_4}, \eqref{eq:lemma_2_5}, and \eqref{eq:lemma_2_6}, respectively};
\EndIf \Comment{End-if at Step\#17}
\Statex {}
\State \textbf{return} the parameter estimates: $\widehat{F}^\ast$, $\widehat{F}(0)$, $\widehat{\zeta}$, $\widehat{C}$, $\widehat{\Gamma}$, and $\widehat{A}$.
\end{algorithmic}
\end{algorithm*}

\normalsize

To this end, we design the procedure in Algorithm \ref{alg:AFAFed_profiling}, whose implementation merits, indeed, three main explicative remarks. First, at each $t$-indexed global iteration, the modified \textit{AFAFed} protocol in the first part of Algorithm \ref{alg:AFAFed_profiling} performs the same steps already reported in the baseline Algorithm \ref{alg:AFAFed_protocol} and, then, it carries out the additional steps detailed in $line\#5$--$line\#8$. These additional steps allow the central server to: (i) store the more recent stochastic gradient communicated by the coworkers (see $line\#6$); and, (ii) compute and store the $t$-th stochastic global gradient $\vv{\widehat{G}}(t)$ (see $line\#8$). Second, after finishing to run the first part of Algorithm \ref{alg:AFAFed_profiling}, central server and coworkers enter the online profiling phase, which is reported in the second part of Algorithm 2 and is carried out in a synchronized way (see the {\sffamily\bfseries parfor} statement at $line\#11$). Third, the vector: $\widehat{\vv{\nabla}F}$ in $line\#15$ is the sample average of the sequence: $\left\{ \vv{\nabla}\widetilde{F}_k(\cdot) \right\}$ of the stochastic local gradients sent to the central server by coworkers during the execution of the first part of Algorithm \ref{alg:AFAFed_profiling}. Hence, $\widehat{\vv{\nabla}F}$ must be properly understood as a noisy estimate of the average value of the full gradient $\vv{\nabla}F$ present in the bounds of \textit{Proposition \ref{prop:1}}. This is, indeed, the main reason why the Eqs. \eqref{eq:lemma_2_4}--\eqref{eq:lemma_2_6} provide only estimates of the (typically, unknown) actual values of the corresponding \textit{AFAFed} parameters.  

Before proceeding, we remark that the reason why the profiling phase of Algorithm \ref{alg:AFAFed_profiling} requires coworker synchronization is that $\zeta$ in Eq. \eqref{eq:zeta_smooth} and $F\!\left( \vv{\overline{w}}(0) \right)$ and $F^\ast$ in Eqs. \eqref{eq:prop_1_2}--\eqref{eq:prop_1_4} are \textit{global} parameters, which characterize the analytical behaviour of the (\emph{a priori} unknown) \textit{global} loss function $F(\cdot)$ in Eq. \eqref{eq:F_global}. For this reason, designing an asynchronous procedure for their online estimates seems to be (if feasible) a challenging task. However, an estimation procedure that does not require inter-coworker synchronization could be designed in the case in which: (i) the global constraint in Eq. \eqref{eq:zeta_smooth} is relaxed and replaced by a set of $K$ per-coworker local constraints, each one depending on a local parameter: $\zeta_k$, $k = 1,\ldots,K$; and, (ii) the global system parameter $F\!\left( \vv{\overline{w}}(0) \right)$ (resp., $F^\ast$) is replaced by its upper (resp., lower) bound: $\max_k \left\{ F_k\!\left( \vv{\overline{w}}(0) \right) \right\}$ (resp., $\min_k \left\{ F_k\!\left( \vv{\overline{w}}(T) \right) \right\}$). The drawback is that the resulting convergence bounds would be somewhat looser than the ones reported in Eqs. \eqref{eq:prop_1_2}--\eqref{eq:prop_1_4}. Finally, we stress that, since Algorithm \ref{alg:AFAFed_profiling} is required \textit{only} for the numerical evaluation of the bounds, it \textit{does not} impact on the \textit{AFAFed} protocol of Algorithm \ref{alg:AFAFed_protocol}, which is \textit{asynchronous}, regardless from the synchronous/asynchronous nature of Algorithm \ref{alg:AFAFed_profiling}.

\subsection{Throughput-maximizing distributed management of stream arrivals}
\label{sec:AFAFed_throughput}

\begin{figure*}[htbp]
\centering
\includegraphics[width=0.85\textwidth]{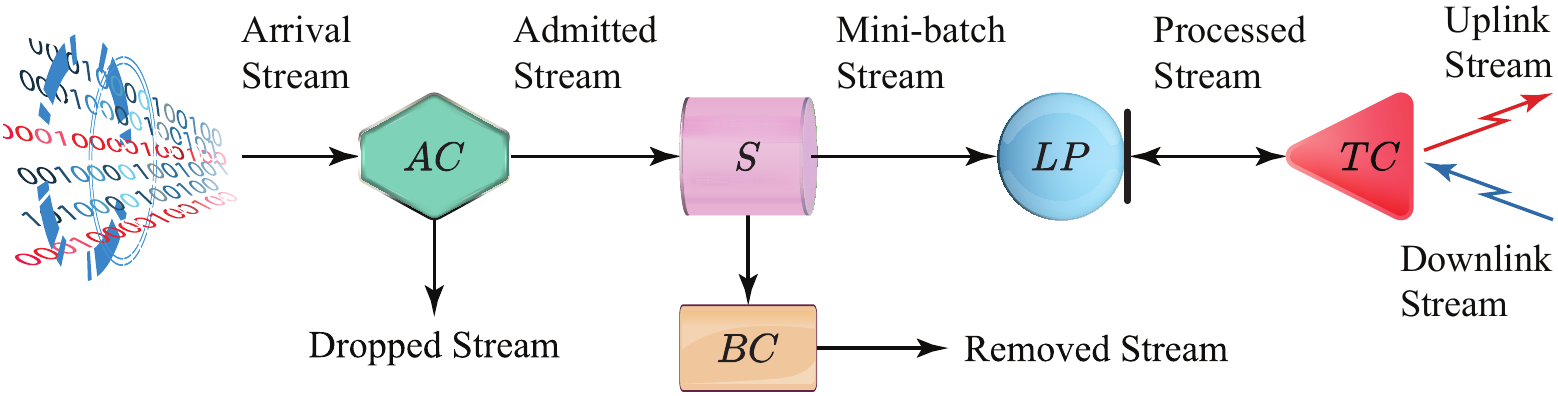}
\caption{Designed functional architecture of an \textit{AFAFed} co-worker. S:= Storage buffer; BC := Buffer Controller; AC := Access Controller; LP := Local Processor; TC := Trans-Ceiver.}
\label{fig:functional}
\end{figure*}

\begin{figure*}[htb]
\centering
\includegraphics[width=0.55\textwidth]{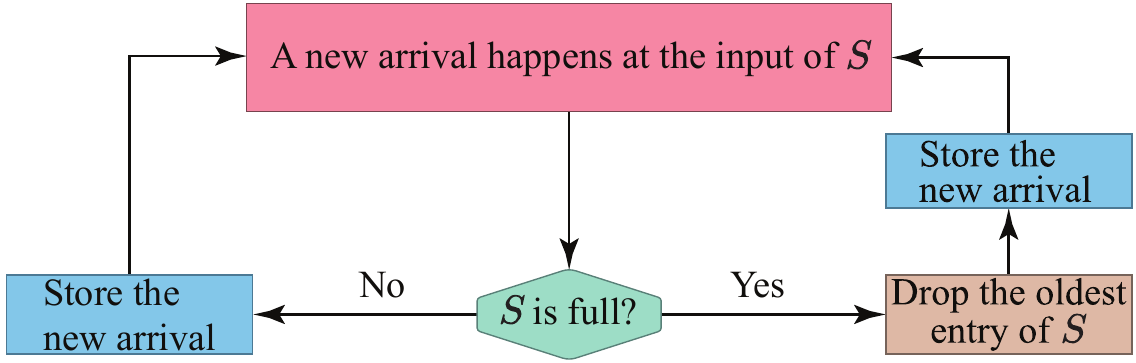}
\caption{Flowchart of the proposed \textit{Access Control Policy} (\textit{ACP}).}
\label{fig:ACP}
\end{figure*}

\begin{figure*}[htb]
\centering
\includegraphics[width=0.5\textwidth]{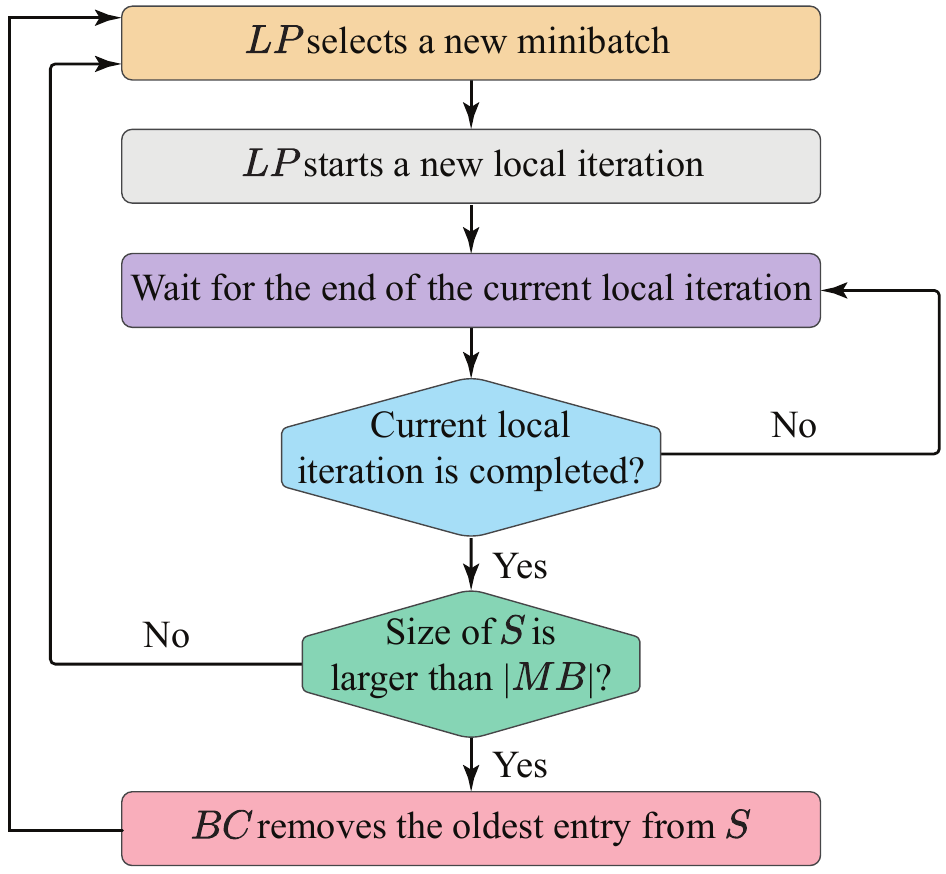}
\caption{Flowchart of the proposed \textit{Buffer Control Policy} (BCP). S:= Storage buffer; BC := Buffer Controller; LP := Local Processor.}
\label{fig:BCP}
\end{figure*}

In the IoT-oriented \textit{AFAFed} ecosystem of Fig.\ \ref{fig:scenario}, coworkers are typically equipped with local sensors, which perform sensing actions on the surrounding environment according to some specified (possibly, energy-conserving) event-driven policy for acquiring new training data. Therefore, since the sensing instants are not known in advance, they should be properly modeled as streams of random arrival times. As a matter of this fact, in order to avoid that the execution of the primal/dual iterations in Eqs. \eqref{eq:SGD_iteration_1} and \eqref{eq:SGD_iteration_2} slows down (or even stalls at all) due to the stretching of the data inter-arrival times, the main guideline pursued in the design of the \textit{AFAFed} stream management policy is to make the execution rate of the local iterations at the coworkers as much as possible \textit{uncoupled} from the corresponding (uncontrollable) arrival rate of new training data. To this end, after designing the coworker functional architecture of Fig.\ \ref{fig:functional}, we propose the distributed stream management policy composed by the \textit{Access Control Policy} (ACP) and \textit{Buffer Control Policy} (BCP) detailed by the flow charts in Figs.\ \ref{fig:ACP} and \ref{fig:BCP}, respectively.

According to Fig.\ \ref{fig:functional}, each \textit{AFAFed} coworker is assumed to be equipped with: (i) a buffer $\mathcal{S}$, which is capable of storing a finite number $|\mathcal{S}| \geq 1$ of training data; (ii) an \textit{Access Controller} (\textit{AC}), which performs selective admission of new training data; (iii) a \textit{Buffer Controller} (\textit{BC}), which manages the current content of buffer $\mathcal{S}$ by removing the oldest stored data; (iv) a \textit{Local Processor} (\textit{LP}), which runs the primal/dual local iterations of Eqs. \eqref{eq:SGD_iteration_1} and \eqref{eq:SGD_iteration_2}, and, at each iteration, builds up a mini-batch of size $|MB| \geq 1$ by randomly sampling data from the current content of buffer $\mathcal{S}$; and, (v) a \textit{Trans-Ceiver} (\textit{TC}), which manages the up/down communications between coworker and central server.

The task of the \textit{AC} is to implement the \textit{ACP} reported in Fig.\ \ref{fig:ACP}. This last guarantees that the number of stored training data does not exceed the (finite) capacity $|\mathcal{S}|$ of the available buffer, while the current content of buffer $\mathcal{S}$ is always composed of the last arrived (i.e. \textit{newest}) $|\mathcal{S}|$ training data.
Task of the \textit{BC} and \textit{LP} of Fig.\ \ref{fig:functional} is to jointly implement the BCP of Fig.\ \ref{fig:BCP}. In this regard, an examination of the corresponding flow chart points out that, by design, the proposed \textit{BCP} dynamically removes the oldest entry of buffer $\mathcal{S}$, but only when the number of currently stored data is \textit{larger} than the mini-batch size $|MB|$. By doing so, the designed \textit{BCP}: (i) dynamically frees space for storing new arrivals, in order to \textit{minimize} the drop rate of the \textit{AC}; while: (ii) guaranteeing that at least $|MB|$ training data are \textit{always} stored by the buffer. 
This last feature is, indeed, crucial to guarantee that the designed \textit{BCP} of Fig.\ \ref{fig:BCP} is computing \textit{throughput-maximizing}. By definition, in our setting, computing \textit{throughput-maximizing} means that the steady-state computing throughput (i.e., the computing rate) at which the local iterations of Eqs. \eqref{eq:SGD_iteration_1} and \eqref{eq:SGD_iteration_2} are run by each coworker may equate the corresponding per-coworker maximum processor's speed, \textit{regardless} from the (random and out-of-control) rate of arrivals of new stream data (see Fig. \ref{fig:functional}). Hence, the expected key benefit arising from the designed \textit{throughput-maximizing} policy is that it would shorten the resulting \textit{AFAFed} average training times, because it allows the coworkers to run their local iterations without suffering from stalling phenomena. To check this property, we observe that, in order to run each local iteration in Eq. \eqref{eq:SGD_iteration_1}, $coworker\#k$ must evaluate the per-minibatch gradient $\vv{\nabla}\widetilde{F}_k(\cdot)$ in Eq. \eqref{eq:DF_tilde_k}. This last evaluation requires, in turn, to perform the average of the underlying local loss function of Eq. \eqref{eq:DF_tilde_k} over a set of $\left| MB_k \right|$ data randomly sampled from the local buffer of $coworker\#k$, and this last operation may be carried out without stalling \textit{only} if the designed \textit{BCP} of Fig.\ \ref{fig:BCP} guarantees that the $k$-th local buffer \textit{always} stores at least $\left| MB_k \right|$ elements.

\section{Conclusion}
\label{sec:conc}

In this paper, we designed and analyzed the convergence properties of \textit{AFAFed}, a novel asynchronous distributed and adaptive scheme for performing FL in heterogeneous stream-oriented IoT scenarios affected by packet-loss communication phenomena. The key feature of \textit{AFAFed} is the implementation of distributed sets of local adaptive tolerance thresholds and global centralized adaptive fairness coefficients, whose symbiotic action allows \textit{AFAFed} to attain right personalization-vs.-fairness trade-offs in learning scenarios featured by coworker heterogeneity, skewed training data, and unreliable communications. \textit{AFAFed} numerical performance evaluation and performance comparisons against FL benchmark schemes are work in progress, and will be published in a fiture work.

\section*{Acknowledgment}
\label{sec:ack}

This work has been supported by the projects: ``DeepFog -- Optimized distributed implementation of Deep Learning models over networked multitier Fog platforms for IoT stream applications'' funded by Sapienza University of Rome Bando 2020 and 2021; and, ``End-to-End Learning for 3D Acoustic Scene Analysis (ELeSA)'' funded by Sapienza University of Rome Bando Acquisizione di medie e grandi attrezzature scientifiche 2018. The work is also partially supported by the research collaboration with the Institute of Informatics and Telematics (IIT-CNR) on the theme: ``Design and software development of Fog Computing platforms for the support of Deep Learning algorithms for the real-time image analysis''

\appendix

\section{Taxonomy and default parameter setting}
\label{appendix:A}

Table \ref{tab:Tab_A1} details the main symbols used in this paper, their meaning/role, and the corresponding default setup.

\footnotesize{
\begin{center}
\rowcolors{2}{gray!25}{gray!5}
\renewcommand{\arraystretch}{1.4}
\begin{longtable}{
m{0.26\textwidth}<{\raggedright}
m{0.42\textwidth}<{\raggedright}
m{0.24\textwidth}<{\raggedright}}
\caption{Main taxonomy and default setup. AF:= AFAFed; FAG:= FedAvg; FPR:= FedProx; CS-SGD:= Cent.Sync.-SGD.}
\label{tab:Tab_A1}\\
%
\specialrule{1pt}{0pc}{0pc}
\rowcolor[HTML]{babfc6}
\multicolumn {1}{c}{\textbf{Parameter}}       & 
\multicolumn {1}{c}{\textbf{Definition}}    & 
\multicolumn {1}{c}{\textbf{Default Settings}} \\[1ex]
\specialrule{0.5pt}{0pc}{0pc}
$K$        & Number of involved coworkers    & 100 \\[1ex] 
$T$        & Number of aggregations performed by the central server & Application dependent \\[1ex] 
%
%
$t^{(k)}$ (resp., $t$) & Integer-valued non-negative dimensionless $k$-th local time index (resp., global time index)  & Application dependent \\[1ex] 
$\ell(\cdot)$  & Loss function  & Application dependent \\[1ex] 
$F(\cdot)$ and $\vv{\nabla}F(\cdot)$  & Global loss function and related vector gradient & Application dependent \\[1ex] 
$\widetilde{F}_k(\cdot)$ and $\vv{\nabla}\widetilde{F}_k(\cdot)$  & Local $k$-th empirical risk function and related vector gradient & Application dependent \\[1ex] 
$\left\{ \lambda_k \right\}$  & Set of the (dimensionless non-negative) fairness coefficients & Adaptively tuned \\[1ex] 
$\left\{ \mu_k \right\}$ and $\left\{ \overline{\mu}_k \right\}$ &  Set of the (dimensionless non-negative) local Lagrange multipliers and their time averages & Adaptively tuned \\[1ex] 
$\left\{ B_k \right\}$  & Set of the (dimensionless non-negative) local tolerance thresholds & Adaptively tuned \\[1ex] 
$\left\{ Iter_k \right\}$  & Set of the (integer-valued positive) numbers of consecutive local model updatings & Adaptively tuned \\[1ex] 
$\left\{ S_k \right\}$  & Spectrum of the local training data sets & Application dependent \\[1ex] 
$\left\{ MB_k \right\}$ and $\left\{\left| MB_k \right|\right\}$  & Spectra of local minibatches and their sizes & $\left| MB_k \right| =$ 16, 32 \\[1ex] 
$\left\{ \Sigma_k \right\}$ and $\overline{\Sigma^2}$  & Spectrum of the local gradient variances and its corresponding ensemble average & Application dependent \\[1ex] 
$\left\{ P_k \right\}$ and $\left\{ \xi_k \right\}$  & Spectra of the access probabilities and related binary-valued indicator functions & See Eqs. \eqref{eq:P_k} and \eqref{eq:csi_k} \\[1ex] 
$B_0$, $de$, and $\gamma$  & Non-negative dimensionless reference tolerance threshold, decay exponent and power exponent & $B_0 = 1.0$, $de \in \left[ 0, \, 0.6 \right]$, and $\gamma = 0.1$ \\[1ex] 
$\widetilde{\mu}$ and $\sigma_{\tilde{\mu}}^2$  & Non-negative dimensionless ensemble-averaged Lagrange multiplier and corresponding variance at the central server & Adaptively tuned \\[1ex] 
$\overline{I}$ (resp., $\overline{I^{2}}$)  & Non-negative dimensionless ensemble-average number (resp., squared number) of per-coworker consecutive local model updatings & Adaptively tuned \\[1ex] 
$age(t)$  & Non-negative integer-valued age of the $t$-th aggregation at the central server & Application dependent \\[1ex] 
$C$, $\Gamma$, and $A$  & Dimensionless non-negative system parameters & Profiled online as in Eqs. \eqref{eq:lemma_2_4}, \eqref{eq:lemma_2_5}, and \eqref{eq:lemma_2_6} \\[1ex] 
$\beta(t)$  & Dimensionless non-negative mixing parameter for model aggregation & Adaptively tuned as in Eq. \eqref{eq:beta_t} \\[1ex] 
$\left\{ \eta_k^{(0)} \right\}$ and $\left\{ \eta_k^{(1)} \right\}$  & Sets of the local primal and dual step-sizes & Adaptively tuned as in Eqs. \eqref{eq:eta_update_1} and \eqref{eq:eta_update_2} \\[1ex] 
%
%
%
%
%
%
$Cat\#i$, $i=1,2,3$  & Set of coworkers of $i$-th category & $\left|Cat\#1\right| = \left|Cat\#3\right| = 30$ and $\left|Cat\#2\right| = 40$  \\[1ex] 
$\left\{ s_k \right\}$  & Set of coworker processing speeds (in (CPU cycles/round)) & $50 \times 10^7$, $25 \times 10^7$, and $1.0 \times 10^7$ (CPU cycles/round) for coworkers of $Cat\#1$, \#2 and \#3, respectively  \\[1ex] 
$\left\{ R_k \right\}$  & Set of coworker communication rates in uplink (in (bit/round)) & $100 \times 10^4$, $50 \times 10^4$, and $2 \times 10^4$ (bit/round) for coworkers of $Cat\#1$, \#2 and \#3, respectively  \\[1ex] 
%
%
$\left\{ P_k^{(LOSS)}\right\}$  & Set of the per-uplink packet-loss probabilities &	0.00, 0.25, and 0.50 \\[1ex] 
%
%
$\left\{ \mathcal{D}_k \right\}$  & Spectrum of the probability distributions of the training data sets & IID, 1-nonIID, and 2-nonIID  \\[1ex] 
$\Omega^{(AF)}$, $\Theta^{(AF)}$, $\Psi^{(AF)}$, $V^{(AF)}$, and $\Phi^{(AF)}$  & Designed functions for the adaptive tuning of the \textit{AFAFed} numbers of consecutive local updates, tolerance thresholds, fairness coefficients, and  mixing parameter	& Set as in Eqs. \eqref{eq:Omega_max_log}, \eqref{eq:Theta_example}, \eqref{eq:example_up_down_func}, and \eqref{eq:Phi_age_example} \\[1ex] 
%
%
$Iter_{MAX}^{(AF)}$  & \textit{AFAFed} maximum number of consecutive local updates & 1 and 30 \\[1ex] 
%
%
$I^{(FAG)}$ & \textit{FedAvg} number of consecutive local updates & Statically set to $\left\lceil \overline{I}^{(AF)} \right\rceil$ \\[1ex] 
$I^{(FPR)}$  & \textit{FedProx} number of consecutive local updates & Statically set to $\left\lceil \overline{I}^{(AF)} \right\rceil$ \\[1ex] 
%
%
$\beta^{(FAS)}$  & \textit{FedAsync} mixing parameter & Set to: $1 / \sqrt{1 + age(t)}$ \\[1ex] 
$I^{(FAS)}$  & \textit{FedAsync} number of consecutive local updates & Statically set to $\left\lceil \overline{I}^{(AF)} \right\rceil$ \\[1ex] 
%
%
$s^{(CS-SGD)}$  & \textit{CS-SGD} processing speed (in (CPU cycles/round)) & $3.28 \times 10^{10}$ (CPU cycles/round) \\[1ex] 
$\left| MB^{(CS-SGD)} \right|$  & \textit{CS-SGD} mini-batch size & $K \times \left|MB\right|$ \\[1ex] 
\bottomrule
\end{longtable}
\renewcommand{\arraystretch}{1.0}
\end{center}
}

\normalsize

\section{Derivations of some auxiliary results}
\label{appendix:B}

In this Appendix \ref{appendix:B}, we provide the proofs of some auxiliary results. 

\vspace{1em}\noindent
\textsc{Proof of the bound in Eq. \eqref{eq:E_D_k_var_3} on the gradient variance}.
The derivation of the bound in Eq. \eqref{eq:E_D_k_var_3} relies on the following chain of developments: 

\begin{equation}
\begin{split}
E_{\mathcal{D}_k}\left\{ \left\| \vv{\nabla}\widetilde{F}_k\!\left( \vv{w} \right) - \vv{\nabla}F\!\left( \vv{w} \right) \right\|^2 \right\} &= E_{\mathcal{D}_k}\left\{ \left\| \left( \vv{\nabla}\widetilde{F}_k\!\left( \vv{w} \right) - \vv{\nabla}F_k\!\left( \vv{w} \right) \right) - \left( \vv{\nabla}F\!\left( \vv{w} \right) - \vv{\nabla}F_k\!\left( \vv{w} \right) \right) \right\|^2 \right\}  \\
  & \overset{(a)}{=} E_{\mathcal{D}_k}\left\{ \left\| \vv{\nabla}\widetilde{F}_k\!\left( \vv{w} \right) - \vv{\nabla}F_k\!\left( \vv{w} \right) \right\|^2 + \left\| \vv{\nabla}F_k\!\left( \vv{w} \right) - \vv{\nabla}F\!\left( \vv{w} \right) \right\|^2 \right\} \overset{(b)}{\leq} \frac{\sigma_{k,1}^2}{\left| MB_k \right|} + \sigma_{k,2}^2,
\end{split}
\label{eq:B1}
\end{equation}
where: (a) follows from the assumption of unbiased gradient estimation in Eq. \eqref{eq:E_D_k}; and, (b) stems from the assumption in Eqs. \eqref{eq:E_D_k_var_1} and \eqref{eq:E_D_k_var_2} of bounded gradient variances.
\hfill\qedsymbol

\vspace{1em}\noindent
\textsc{Proof of Lemma \ref{lem:1}}.
In order to prove the bounds on the $C$ parameter in Eq. \eqref{eq:lemma_1_3}, we begin to observe that the following chain of inequalities holds: 
\begin{equation}
\left( E_{\xi \sim P}\left\{ \widehat{\vv{G}}(\cdot) \right\} \right)^T \vv{\nabla}F(\cdot) \overset{(a)}{=} \left( E_{\xi \sim P}\left\{ \vv{\nabla}F(\cdot) + \vv{n} \right\} \right)^T \vv{\nabla}F(\cdot) \overset{(b)}{=} \left\| \vv{\nabla}F(\cdot) \right\|^2 \left( 1 + \gamma_0 \right),
\label{eq:B2}
\end{equation}
where: (a) follows from the definition of the estimation noise: $\widehat{\vv{G}}(\cdot) \equiv \vv{\nabla}F(\cdot) + \vv{n}$; and, (b) stems from Eq. \eqref{eq:lemma_1_1}. Hence, by leveraging Eq. \eqref{eq:B2}, we may re-write the condition in Eq. \eqref{eq:assump_4_1} as in: $\left\| \vv{\nabla}F(\cdot) \right\|^2 \left( 1 + \gamma_0 \right) \geq C \left\| \vv{\nabla}F(\cdot) \right\|^2$, which, in turn, leads to the upper bound: $\left( 1 + \gamma_0 \right) \geq C$. Furthermore, since $C$ must be, by definition, positive, we must have: $\gamma_0 > -1$.

In order to derive the bound in Eq. \eqref{eq:lemma_1_3} on the $\Gamma$ parameter, we observe that the following equality chain holds:
\begin{equation}
\left\| E_{\xi \sim P}\left\{ \widehat{\vv{G}}(\cdot) \right\} \right\| = \left\| \vv{\nabla}F(\cdot) + E_{\xi \sim P}\left\{ \vv{n} \right\} \right\| \overset{(a)}{=} \left\| \vv{\nabla}F(\cdot) + \gamma_0 \vv{\nabla}F(\cdot) \right\| \overset{(b)}{=} \left( 1 + \gamma_0 \right) \vv{\nabla}F(\cdot),
\label{eq:B3}
\end{equation}
where: (a) arises from Eq. \eqref{eq:lemma_1_1}; and, (b) is assured by the fact that $\gamma_0 > -1$. Therefore, after inserting Eq. \eqref{eq:B3} into the l.h.s. of Eq. \eqref{eq:assump_4_2}, we arrive at the lower bound on $\Gamma$ of Eq. \eqref{eq:lemma_1_3}. 

Finally, in order to derive the lower bound on $A$ in Eq. \eqref{eq:lemma_1_3}, we develop the defining relationship of Eq. \eqref{eq:assump_4_3} of $\sigma_{\widehat{\vv{G}}}^2$ as follows:
\begin{equation}
\begin{split}
\sigma_{\widehat{\vv{G}}}^2 &= E_{\xi \sim P}\left\{ \left\| \widehat{\vv{G}}(\cdot) \right\|^2 \right\} - \left\| E_{\xi \sim P}\left\{ \widehat{\vv{G}}(\cdot) \right\} \right\|^2 \overset{(a)}{=} E_{\xi \sim P}\left\{ \left\| \vv{\nabla}F(\cdot) + \vv{n} \right\|^2 \right\} - \left\| E_{\xi \sim P}\left\{ \vv{\nabla}F(\cdot) + \vv{n} \right\} \right\|^2 \\
 & \overset{(b)}{=} \left( 1 + \gamma_1 + 2 \gamma_0 \right) \left\| \vv{\nabla}F(\cdot) \right\|^2 + b_0 - \left( 1 + \gamma_0 \right)^2 \left\| \vv{\nabla}F(\cdot) \right\|^2 \equiv \left( \gamma_1 - \gamma_0^2 \right) \left\| \vv{\nabla}F(\cdot) \right\|^2 + b_0,
\end{split}
\label{eq:B4}
\end{equation}
where: (a) follows from the definition: $\widehat{\vv{G}}(\cdot) = \vv{\nabla}F(\cdot) + \vv{n}$ of the estimation noise; and, (b) stems from Eqs. \eqref{eq:lemma_1_1} and \eqref{eq:lemma_1_2}. Hence, after noting that the condition: $0 \leq \left( \gamma_1 - \gamma_0^2 \right) \leq 1$, in Eq. \eqref{eq:lemma_1_2} guarantees that: (i) the last expression in Eq. \eqref{eq:B4} is non-negative for non-negative $b_0$; and, (ii) the factor $\left( \gamma_1 - \gamma_0^2 \right)$ in Eq. \eqref{eq:B4} is upper bounded by the unit, the following upper bound follows from Eq. \eqref{eq:B4}:
\begin{equation}
\sigma_{\widehat{\vv{G}}}^2 \leq \left\| \vv{\nabla}F(\cdot) \right\|^2 + b_0.
\label{eq:B5}
\end{equation}
Therefore, the comparison of the bounds in Eqs. \eqref{eq:assump_4_4} and \eqref{eq:B5} proves the validity of the lower bound on $A$ in Eq. \eqref{eq:lemma_1_3}. Finally, we note that, in the case in which the stochastic gradient $\widehat{\vv{G}}(\cdot)$ would provide an unbiased estimate of the full gradient $\vv{\nabla}F(\cdot)$, the expectation of the corresponding estimation noise $\vv{n}$ vanishes and its second-order moment must be constant. This requires, in turn, that both $\gamma_0$ and $\gamma_1$ in Eqs. \eqref{eq:lemma_1_1} and \eqref{eq:lemma_1_2} must vanish. The proof of Lemma \ref{lem:1} is now completed. 
\hfill\qedsymbol

\vspace{1em}\noindent
\textsc{Derivation of the scaling relationships in Eqs.} (68a), (68b), \textsc{and} (68c). 
We begin to rewrite the primal iteration of Eq. \eqref{eq:SGD_iteration_1} in the following compact form:
\begin{equation}
\vv{w}_k\!\left( t^{(k)} + 1 \right) = \vv{w}_k\!\left( t^{(k)} \right) - \eta_k^{(0)} \vv{\nabla} Z_k\!\left( \vv{w}_k\left( t^{(k)} \right) \right), \quad t^{(k)} = 0, 1, \ldots,
\label{eq:B6}
\end{equation}
with the dummy position: 
\begin{equation}
\vv{\nabla} Z_k\!\left( \vv{w}_k\left( t^{(k)} \right) \right) \triangleq \lambda_k^{(LAST)} \vv{\nabla}\widetilde{F}_k\!\left( \vv{w}_k\!\left( t^{(k)} \right) \right) + \mu_k\!\left( t^{(k)} \right) \left( \vv{w}_k\left( t^{(k)} \right) - \vv{\overline{w}}_k^{(LAST)} \right).
\label{eq:B7}
\end{equation}
Afterwards, without loss of generality, let us assume that $coworker\#k$: (i) starts a cluster of $Iter_k \geq 1$ iterations at its local index time $t^{(k)}$ by moving from $\vv{\overline{w}}_k^{(LAST)}$; and, then, (ii) communicates its updated local model: $\vv{w}_k\!\left( t^{(k)} + Iter_k \right)$ to the central server at the global time index: $t+1$. Hence, by iterating Eq. \eqref{eq:B6} from: $t^{(k)}$ up to: $t^{(k)} + Iter_k$, we obtain: 
\begin{equation}
\vv{w}_k\!\left( t^{(k)} + Iter_k \right) = \vv{w}_k^{(LAST)} - \sum_{j=t^{(k)}}^{t^{(k)} + Iter_k - 1} \eta_k^{(0)}(j) \, \vv{\nabla} Z_k\!\left( \vv{w}_k(j) \right).
\label{eq:B8}
\end{equation}
Since, by assumption, the local time index: $t^{(k)} + Iter_k$ at which $coworker\#k$ ends its current iteration cluster equates to the global time index $(t + 1)$ at which it communicates its updated model to the central server (i.e., $t^{(k)} + Iter_k \equiv t + 1$, and $\xi_k(t+1) \equiv 1$), the resulting expression assumed by the global stochastic gradient $\widehat{\vv{G}}(\cdot)$ may be obtained by inserting Eq. \eqref{eq:B8} into the defining relationship of Eq. \eqref{eq:G} , and, then, it reads as follows:
\begin{equation}
\widehat{\vv{G}}\!\left( \vv{\overline{w}}(t); \xi_k(t+1) \equiv 1 \right) = \vv{\overline{w}}(t) - \vv{w}_k^{(LAST)} + \sum_{j=t^{(k)}}^{t^{(k)} + Iter_k\left(t^{(k)}\right) - 1} \eta_k^{(0)}(j) \, \vv{\nabla} Z_k\!\left( \vv{w}_k(j) \right).
\label{eq:B9}
\end{equation}
Hence, after performing the expectation of both sides of Eq. \eqref{eq:B9} over the set of the access probabilities in Eq. \eqref{eq:P_k}, we arrive (after some tedious but quite standard algebras) at the following scaling relationships for the first and second-order moments of $\widehat{\vv{G}}(\cdot)$: 
\begin{equation}
E_{\xi \sim P}\left\{ \widehat{\vv{G}}(\cdot) \right\} \propto \mathcal{O}\!\left( \overline{I} \right),
\label{eq:B10}
\end{equation}
\begin{equation}
\mathcal{O}\!\left( \overline{I}^2 \right) \leq E_{\xi \sim P}\left\{ \left\| \widehat{\vv{G}}(\cdot) \right\|^2 \right\} \leq \mathcal{O}\!\left( \overline{I^2} \right),
\label{eq:B11}
\end{equation}
and,
\begin{equation}
\mathcal{O}\!\left( E_{\xi \sim P}\left\{ \left\| \widehat{\vv{G}}(\cdot) \right\|^2 \right\} \right) \leq \mathcal{O}\!\left( \overline{\Sigma^2} \right).
\label{eq:B12}
\end{equation}
Finally, after observing that, from Eqs. \eqref{eq:B11} and \eqref{eq:B12}, the variance $\sigma_{\widehat{\vv{G}}}^2$ in Eq. \eqref{eq:assump_4_3} scales (at most) as the product $\mathcal{O}\left( \overline{I^2} \: \overline{\Sigma^2} \right)$, the relationships in Eqs. \eqref{eq:bound_scale_1}, \eqref{eq:bound_scale_2}, and \eqref{eq:bound_scale_3} follow by accounting for the scaling expressions in Eqs. \eqref{eq:B10}--\eqref{eq:B12} in the bounds of Eqs. \eqref{eq:assump_4_1}, \eqref{eq:assump_4_2}, and \eqref{eq:assump_4_4}.      
\vspace{1em}\hfill\qedsymbol

\vspace{1em}\noindent
\textsc{Proof of Lemma \ref{lem:2}}.
From the bound in Eq. \eqref{eq:assump_4_1} on $C$, we obtain: $C \leq \left( E_{\xi \sim P}\left\{ \widehat{\vv{G}} \right\} \right)^T \, \vv{\nabla}F / \left\| \vv{\nabla}F \right\|^2$, so that, since we must have $C > 0$, we conclude that the condition in Eq. \eqref{eq:lemma_2_1} is both necessary and sufficient for the feasibility of the estimate in Eq. \eqref{eq:lemma_2_4}. Regarding the parameter $\Gamma$ in Eq. \eqref{eq:assump_4_1}, from \textit{Assumption 4.1}, we must have: $\Gamma > C$, so that without loss of generality, we may pose:
\begin{equation}
\Gamma = \left( 1 + k_0 \right) C,
\label{eq:B13}
\end{equation}
with $k_0 \geq 0$. Furthermore, from the bound in Eq. \eqref{eq:assump_4_1}, we obtain the following limit value on the feasible values of $\Gamma$: 
\begin{equation}
\Gamma \geq \frac{\left\| E_{\xi \sim P}\left\{ \widehat{\vv{G}} \right\} \right\|}{\left\|\vv{\nabla}F\right\|}.
\label{eq:B14}
\end{equation}
Hence, a comparison of the relationships in Eqs. \eqref{eq:B13} and \eqref{eq:B14} directly leads to the conclusion that $\widehat{\Gamma}$ in Eq. \eqref{eq:lemma_2_5} provides a feasible estimate of $\Gamma$ if and only if the condition in Eq. \eqref{eq:lemma_2_2} is met. Finally, the bound in Eq. \eqref{eq:assump_4_4} may be rewritten as: $A \geq \sigma_{\widehat{\vv{G}}}^2 - \left\| \vv{\nabla}F\!\left( \vv{\overline{w}} \right) \right\|^2$, with $\sigma_{\widehat{\vv{G}}}^2$ defined as in Eq. \eqref{eq:assump_4_3}. Hence, since $A$ must be non-negative by \textit{Assumption 4.2}, we conclude that $\widehat{A}$ in Eq. \eqref{eq:lemma_2_6} is a feasible estimate of $A$ if and only if the condition in Eq. \eqref{eq:lemma_2_3} is met.
\hfill\qedsymbol

\section{Proof of Proposition \ref{prop:1}}
\label{appendix:C}

\textit{Assumption 3.2} on the $\zeta$-smoothness of the local loss functions guarantees that also the resulting global loss function $F(\cdot)$ in Eq. \eqref{eq:F_global} is $\zeta$-smooth (see, for example, Lemma 1 of \cite{Wang2019}), so that, from Lemma 3.4 of \cite{Bubeck2015}, it meets the following inequality:
\begin{equation}
F\!\left( \vv{w}_1 \right) \leq F\!\left( \vv{w}_2 \right) + \vv{\nabla}F\!\left( \vv{w}_2 \right)^T \left( \vv{w}_1 - \vv{w}_2 \right) + \frac{\zeta}{2} \left\| \vv{w}_1 - \vv{w}_2 \right\|^2, \quad \forall \vv{w}_1, \vv{w}_2 \in \mathbb{R}^l.
\label{eq:C1}
\end{equation}
Hence, after posing: $\vv{w}_1 = \vv{\overline{w}}(t+1)$, and: $\vv{w}_2 = \vv{\overline{w}}(t)$ in Eq. \eqref{eq:C1}, and, then, replacing $\vv{\overline{w}}(t+1)$ by the r.h.s. of Eq. \eqref{eq:CS_update_2_2}, we obtain:
\begin{equation}
F\!\left( \vv{\overline{w}}(t+1) \right) - F\!\left( \vv{\overline{w}}(t) \right) \leq - \beta(t) \vv{\nabla}F\!\left( \vv{\overline{w}}(t) \right)^T \, \widehat{\vv{G}}\!\left( \vv{\overline{w}}(t); \left\{ \xi_k(t + 1) \right\} \right) + \frac{\zeta}{2} \beta(t)^2 \left\| \widehat{\vv{G}}\!\left( \vv{\overline{w}}(t); \left\{ \xi_k(t + 1) \right\} \right) \right\|^2,
\label{eq:C2}
\end{equation}
which, by performing the expectation of both sides of Eq. \eqref{eq:C2} the access probabilities of Eq. \eqref{eq:P_k} and, then, exploiting the gradient coherence assumption of Eq. \eqref{eq:assump_4_1}, leads, in turn, to:
\begin{equation}
E_{\xi \sim P}\left\{ F\!\left( \vv{\overline{w}}(t+1) \right) - F\!\left( \vv{\overline{w}}(t) \right) \right\} \leq - \beta(t) C \left\| \vv{\nabla}F\!\left( \vv{\overline{w}}(t) \right) \right\|^2 + \frac{\zeta}{2} \beta(t)^2 E_{\xi \sim P}\left\{ \left\| \widehat{\vv{G}}\!\left( \vv{\overline{w}}(t); \left\{ \xi_k(t + 1) \right\} \right) \right\|^2 \right\}.
\label{eq:C3}
\end{equation}
Furthermore, since the introduction of the inequalities in Eqs. \eqref{eq:assump_4_2} and \eqref{eq:assump_4_4} into the defining relationship of Eq. \eqref{eq:assump_4_3} allows us to upper limit the last expectation in Eq. \eqref{eq:C3} as in:
\begin{equation}
E_{\xi \sim P}\left\{ \left\| \widehat{\vv{G}}\!\left( \vv{\overline{w}}(t); \left\{ \xi_k(t + 1) \right\} \right) \right\|^2 \right\} \leq A + \left( 1 + \Gamma^2 \right) \left\| \vv{\nabla}F\!\left( \vv{\overline{w}}(t) \right) \right\|^2,
\label{eq:C4}
\end{equation}
we may further rework the bound in \eqref{eq:C3} as follows:
\begin{equation}
E_{\xi \sim P}\left\{ F\!\left( \vv{\overline{w}}(t+1) \right) - F\!\left( \vv{\overline{w}}(t) \right) \right\} \leq - \beta(t) \left\| \vv{\nabla}F\!\left( \vv{\overline{w}}(t) \right) \right\|^2 \left[ C - \frac{\zeta}{2} \beta(t) \left( 1 + \Gamma^2 \right) \right] + \frac{\zeta}{2} \beta(t)^2 A.
\label{eq:C5}
\end{equation}
Our next task is to upper bound $\beta(t)$ by a suitable constant, in order to guarantee that the term into the square brackets of Eq. \eqref{eq:C5} is positive. To this end, we enforce the following upper bound:
\begin{equation}
\frac{\zeta}{2} \beta(t) \left( 1 + \Gamma^2 \right) \leq \varepsilon C,
\label{eq:C6}
\end{equation}
where $\varepsilon$ is an user-defined constant, which must fall into the open interval $(0, \, 1)$. So doing, we have that: (i) the term into bracket in Eq. \eqref{eq:C5} is guaranteed to be positive, while the mixing parameter $\beta(t)$ in Eq. \eqref{eq:C6} is \textit{not} forced to vanish; and, (ii) $\beta(t)$ is upper limited as in Eq. \eqref{eq:prop_1_1}, so that the bound in Eq. \eqref{eq:C5} may be reworked as in:
\begin{equation}
E_{\xi \sim P}\left\{ F\!\left( \vv{\overline{w}}(t+1) \right) - F\!\left( \vv{\overline{w}}(t) \right) \right\} \leq - \beta(t) C \left( 1 - \varepsilon \right) \left\| \vv{\nabla}F\!\left( \vv{\overline{w}}(t) \right) \right\|^2 + \frac{\zeta}{2} \beta(t)^2 A.
\label{eq:C7}
\end{equation}
Hence, after performing the telescopic sum of the terms at both sides of Eq. \eqref{eq:C7} over the interval $t \in \left\{ 0, 1, \ldots, T \right\}$, and, then, noting that (see Eq. \eqref{eq:F_star}): $E_{\xi \sim P}\left\{ F\!\left( \vv{\overline{w}}(T) \right) \right\} \geq F^\ast$, we derive the following bound:
\begin{equation}
F^\ast - E_{\xi \sim P}\left\{ F\!\left( \vv{\overline{w}}(0) \right) \right\} \leq C \left( 1 - \varepsilon \right) \sum_{t=0}^{T-1} \beta(t) \left\| \vv{\nabla}F\!\left( \vv{\overline{w}}(t) \right) \right\|^2 + \frac{\zeta}{2} A \sum_{t=0}^{T-1} \beta(t)^2 .
\label{eq:C8}
\end{equation}
So, after dividing both sides of Eq. \eqref{eq:C8} by $T$ and noting that the term: $(1-\varepsilon)$ is positive by design, we directly arrive at the bound in Eq. \eqref{eq:prop_1_2}.

Finally, the bound in Eq. \eqref{eq:prop_1_3} stems from the one in Eq. \eqref{eq:prop_1_2} by accounting for the following two inequalities:
\begin{equation}
E\!\left\{ \frac{1}{T} \sum_{t=0}^{T-1} \beta(t) \left\| \vv{\nabla}F\!\left( \vv{\overline{w}}(t) \right) \right\|^2 \right\} \geq \beta_{MIN} \, E\!\left\{ \frac{1}{T} \sum_{t=0}^{T-1} \left\| \vv{\nabla}F\!\left( \vv{\overline{w}}(t) \right) \right\|^2 \right\},
\label{eq:C9}
\end{equation}
and,
\begin{equation}
E\!\left\{ \frac{1}{T} \sum_{t=0}^{T-1} \beta(t) \right\} \leq E\!\left\{ \frac{1}{T} \sum_{t=0}^{T-1} \beta_{MAX} \right\} \equiv \beta_{MAX},
\label{eq:C10}
\end{equation}
while the last bound in Eq. \eqref{eq:prop_1_4} is obtained by setting $\beta_{MIN} = \beta_{MAX} \equiv \beta$ at the r.h.s. of Eq. \eqref{eq:prop_1_3}. The proof of Proposition \ref{prop:1} is now completed.





\end{document}